\documentclass[5p,times,twocolumn]{elsarticle}
\usepackage{lineno,hyperref}
\usepackage{subcaption}
\usepackage{tikz}
\usepackage{pgfplotstable}
\usepackage{pgfplots}
\usepackage{multirow}
\usepackage{float}
\usepackage{textcomp}
\usepackage{amsmath,amsfonts,amssymb,bbm}	
\usepackage{multicol}
\usepackage{graphicx}
\usepackage{caption}
\usepackage{booktabs}

\usepackage{algorithm}
\usepackage{algpseudocode}
\usepackage{floatflt}
\usepackage{soul}

\usepackage{pifont}
\newcommand{\cmark}{\ding{51}}%
\newcommand{\xmark}{\ding{55}}%

\usepackage[section]{placeins}

\usepackage{xcolor}

\modulolinenumbers[5]

\makeatletter
\def\ps@pprintTitle{%
    \let\@oddhead\@empty
    \let\@evenhead\@empty
    \def\@oddfoot{\reset@font\hfil\thepage\hfil}
    \let\@evenfoot\@oddfoot
}
\makeatother

\begin{document}

\begin{frontmatter}

\title{PULASki: Learning inter-rater variability using statistical distances to improve probabilistic segmentation}

\author[1,2]{Soumick Chatterjee\corref{equalcontribution}}
\author[3]{Franziska Gaidzik\corref{equalcontribution}}
\author[4,5]{Alessandro Sciarra}
\author[4,6,7]{Hendrik Mattern}
\author[3]{G{\'a}bor Janiga}
\author[4,6,7]{Oliver Speck}
\author[1,6]{Andreas N{\"u}rnberger}
\author[8]{Sahani Pathiraja}

\cortext[equalcontribution]{S. Chatterjee and F. Gaidzik have Equal Contribution}

\address[1]{Data and Knowledge Engineering Group, Faculty of Computer Science, Otto von Guericke University Magdeburg, Germany}
\address[2]{Genomics Research Centre, Human Technopole, Milan, Italy}
\address[3]{Laboratory of Fluid Dynamics and Technical Flows, Otto von Guericke University Magdeburg, Germany}
\address[4]{Department of Biomedical Magnetic Resonance, Faculty of Natural Sciences, Otto von Guericke University Magdeburg, Germany}
\address[5]{MedDigit, Department of Neurology, Medical Faculty, University Hospital, Magdeburg, Germany}
\address[6]{German Centre for Neurodegenerative Disease, Magdeburg, Germany}
\address[7]{Centre for Behavioural Brain Sciences, Magdeburg, Germany}
\address[8]{School of Mathematics \& Statistics, UNSW Sydney and UNSW Data Science Hub}

\begin{abstract}
In the domain of medical imaging, many supervised learning based methods for segmentation face several challenges such as high variability in annotations from multiple experts, paucity of labelled data and class imbalanced datasets.  These issues may result in segmentations that lack the requisite precision for clinical analysis and can be misleadingly overconfident without associated uncertainty quantification.  
This work proposes the PULASki method as a computationally efficient generative tool for biomedical image segmentation that accurately captures variability in expert annotations, even in small datasets. This approach makes use of an improved loss function based on statistical distances in a conditional variational autoencoder structure (Probabilistic UNet)
, which improves learning of the conditional decoder compared to the standard cross-entropy particularly in class imbalanced problems.
The proposed method was analysed for two structurally different segmentation tasks (intracranial vessel and multiple sclerosis (MS) lesion) and compare our results to four well-established baselines in terms of quantitative metrics and qualitative output. 
These experiments involve class-imbalanced datasets characterised by challenging features, including suboptimal signal-to-noise ratios and high ambiguity.
 Empirical results demonstrate the PULASKi method outperforms all baselines at the 5\% significance level.  Our experiments are also of the first to present a comparative study of the computationally feasible segmentation of complex geometries using 3D patches and the traditional use of 2D slices.  The generated segmentations are shown to be much more anatomically plausible than in the 2D case, particularly for the vessel task.  Our method can also be applied to a wide range of multi-label segmentation tasks and and is useful for downstream tasks such as hemodynamic modelling (computational fluid dynamics and data assimilation), clinical decision making, and treatment planning.
 
\end{abstract}


\end{frontmatter}

\section{Introduction} 


The recent explosion in deep learning based medical image segmentation techniques has led to a dramatic improvement in the rapid diagnosis, treatment planning and modelling of various diseases \cite{li2021multiple, ni2020global}. Medical image segmentation has specific challenges compared to other classical computer vision tasks, which has spurred the development of several bespoke machine learning tools \cite{chatterjee2022strega,antonelli2022medical,qureshi2023medical}.  Such issues include 1) limited availability of labelled training data, due to the expertise and excessive time required to create annotations; 2) class imbalance, particularly where the class of interest (e.g., tumour) is infrequently represented in the dataset; 3) complex and ambiguous features which are difficult to identify from Magnetic Resonance Imaging (MRI) data with low signal to noise ratio, leading to variation in annotations from different graders for the same image; and 4) considerable anatomic variability between patients.  As a result, quantifying uncertainty in predicted segmentations, regardless of the chosen method, is paramount to prevent over-confident and biased diagnoses \cite{czolbe2021segmentation,varoquaux2022machine}.  Generating multiple plausible segmentations also supports Monte-Carlo based uncertainty quantification in biomedical computational fluid dynamics (CFD) modelling (e.g., \cite{gaidzik2021hemodynamic, chen2020deep}). CFD based modelling of the onset and progression of cardiovascular and neurovascular diseases such as aneurysms requires capturing uncertainty in various components of the modelling procedure, so as to not produce over-confident predictions. This work proposes an accurate and computationally feasible method for generating multiple plausible segmentations from a single image in medical applications plagued by the aforementioned issues.    

Significant research effort to date has focused on improving the accuracy of deep-learning based medical image segmentation, e.g. using convolutional neural networks (CNNs).  By now, the U-Net \cite{Ronneberger2015} and Attention U-Net \cite{oktay2018attention} have emerged as empirically successful methods for image segmentation in a range of computer vision tasks not limited to medical imaging. 
Enhancements of such methods have been developed to remain competitive when trained on small data sets, e.g., via  deformation consistent, multiscale supervision \cite{Chatterjee2020}, along with improved loss functions for heavily class imbalanced data sets, such as the Focal Tversky Loss \cite{abraham2019novel}.  
However, uncertainty quantification of such outputs remains a considerable challenge, which is exacerbated when training data are sparse.  Furthermore, the logits or activations produced from most deep learning based segmentation methods provide only a pixel-wise measure of uncertainty and cannot be reliably used to generate samples from the underlying joint distribution \cite{Monteiro2020,krygier2021quantifying}.  This necessitates tailor-made methods that carefully account for the joint probability between pixels, as in, for instance, Deep Ensembles \cite{caldeira2020deeply} and Stochastic Segmentation Networks \cite{Monteiro2020}. 

One of the most thoroughly investigated methods for uncertainty quantification in deep learning is with the Bayesian framework.  In the Bayesian approach, prior knowledge of one or more components of the architecture (e.g. neural network weights) are combined with data via Bayes theorem.  While Bayesian methods \cite{li2021deep,abdar2021review,kwon2020} are supported by rich theory for quantifying both aleatoric and epistemic uncertainty, it is necessary to resort to approximations of both the prior and posterior distributions in most high dimensional medical imaging applications.  Arguably the most prominent, flexible and easily implementable approach is Monte Carlo dropout \cite{Gal2016, Mobiny2021}, which involves randomly removing elements of the input tensor \cite{hinton2012improving} or removing channels (i.e., feature maps) while using spatial dropouts with convolutional layers \cite{tompson2015efficient}, during both training and inference.  This serves as a fast but somewhat crude approximation to the posterior.  In Bayesian neural networks, weights are treated as random variables whose distributional properties must also be estimated from data.  This is particularly challenging in medical imaging problems where annotated data sets are limited, thus exacerbating the issue of parameter identifiability \cite{khemakhem2020variational}. 
More recently, Variational inference has emerged as a competitive tool for fast approximate Bayesian computations. Variational autoencoder and variational inference methods, such as the Probabilistic U-Net \cite{kohl2018} which is a major focus of this paper, Variational Inference U-Net (VI U-Net) and Multi-Head VI U-Net \cite{Fuchs2022}) employ a simpler parametric distribution to approximate the posterior (e.g. Normal) to drastically simplify computations and typically also capture uncertainty through a lower dimensional latent variable which is more identifiable.  However, these methods rely on a cross-entropy style reconstruction term in the evidence lower bound (ELBO), which is known to be problematic for class imbalanced data sets \cite{abraham2019novel,rezaei2020addressing,tian2022striking}, as is common in medical image segmentation. This paper proposes a modification of this loss function to improve generation of plausible segmentations for class imbalanced datasets with considerable inter-rater variability.   

Another important consideration in generating segmentations, particularly in medical applications where images contain complex intricate details, is whether to use 2D slices or 3D volumes.  When generating 3D volumes, it is common to use smaller 3D patches when a full 3D model is computationally infeasible due to resource limitations. While 3D patches typically provide superior encoding of spatial relationships, they come with increased computational complexity. Despite the clear benefits of 3D models in capturing the spatial structures, there is a notable gap in the literature regarding the use of 3D implementations for generative models, and direct comparisons between 2D and 3D models remain rare. A second goal of this paper is to address this gap by investigating how training on 2D slices compares to training on 3D patches, offering valuable insights into which approach may be more effective, particularly in medical imaging applications.

\subsection{Contributions}

This paper introduces the PULASki method, a computationally efficient generative tool for biomedical image segmentation that effectively captures variability in expert annotations, even within small datasets. Our method relies on an easily implementable and interpretable modification of popular probabilistic U-nets \cite{kohl2018} primarily via improved loss functions.    These loss functions are based on statistical distances, thereby enhancing the learning process of well-established probabilistic U-Nets by incorporating the complete set of annotations from multiple raters simultaneously, rather than learning from a single random annotation at a time (see Sec. \ref{sec:pulaski}). Incorporating statistical distance measures into loss functions can enhance model performance on imbalanced datasets by enabling a more nuanced comparison between predicted and actual data distributions. We consider a range of distances that improve the model's ability to accurately represent minority classes (and local geometric features), while also allowing the model to focus on the underlying data distribution, leading to better generalisation and performance in scenarios characterised by class imbalance. The advantage of our method is its potential to be easily taken up by practitioners given its structural similarity to the U-net and its popularity for image segmentation.  A major focus here is in testing our method on datasets characterising real-world challenges such as: small sample sizes, class-imbalance, high ambiguity and memory requirements for 3D volumes.  We evaluate PULASki on intracranial vessel and multiple sclerosis lesion segmentation tasks, demonstrating superior performance compared to established baselines.  The features of interest are complex in their geometry and ambiguous and there exists appreciable variability in annotations between graders. Furthermore, the paper presents a comparative study of 2D and 3D segmentation models. Although the experiments focused on two specific segmentation tasks, the method is applicable to other segmentation tasks characterised by high inter-rater disagreement. The versatility of the PULASki method has the potential to be extended to various multi-label segmentation tasks, supporting applications in clinical decision-making, treatment planning, and haemodynamic modelling to better understand cardiovascular and neurovascular diseases which are leading causes of death worldwide \cite{fan2018neurovascular,campbell2019ischaemic,chen2020deep}.

\section{Background} 
\label{sec:background}

\begin{figure*}
    \centering
  \includegraphics[width=14cm,height=9cm]{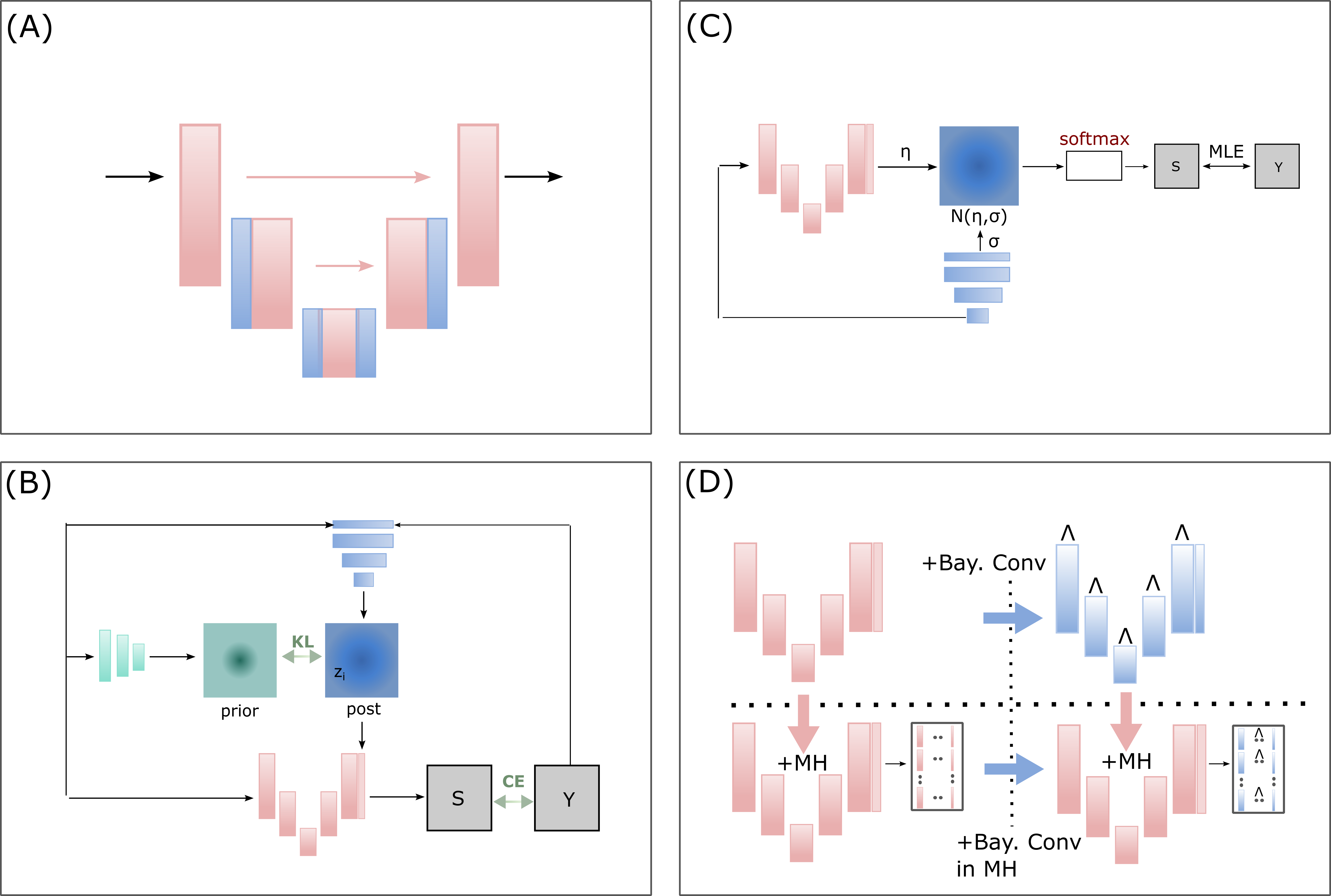}
  \caption{Schematic of Baseline Methods: (A): Monte-Carlo-Dropout (MC-DO); (B): Probabilistic U-Net (ProbU-Net); (C): Stochastic Segmentation Network (SSN); (D): Multi-head Variational Inference U-Net (MH VI U-Net)}
  \label{fig:baselines}
\end{figure*}

\subsection{Notation}
Throughout the manuscript, we make use of the following notation:
\begin{itemize}
    \item [] $x = $  random variable representing image data, i.e. a 2D or 3D matrix containing brightness values for each pixel/voxel.  
    \item [] ${\bf x} = $ dataset consisting of $N$ independent and identically distributed data of the random variable $x$.
    \item [] $x_{ijk}$ = entry in the $i$th position of the first dimension (i.e. row), $j$th position of the second dimension (i.e. column) and $k$th position of the third dimension of a 3D matrix $x$. 
    \item [] $y$ = random variable denoting the ``ground truth'' segmentation of the image $x$.  
    \item [] $c_{ij}$ = categorical random variable of the class for the $ij$th pixel 
    \item [] $s = $ predicted plausible segmentation for an image $x$, with the same dimensions as $x$.  
    \item [] $\{v^i\}_{i=1:M}$ denotes a set of $M$ plausible samples of a random variable $v$. 
    \item [] $z \in \mathbb{R}^p = $ $p$-channel feature map of image-segmentation pair. 
    \item [] $\eta$ = logits i.e. outputs from the final activation layer of a given neural network.  When $x$ is a 2D image of size $d \times d$, $\eta \in \mathbb{R}^{d \times d \times K}$ where $K$ is the desired number of classes. 
    \item [] $M$ = number of samples from posterior net or no. of plausible segmentations per image.  
\end{itemize}

Finally, we use $'.'$ to indicate the multiplication operation and reserve use of $\times$ when defining sizes of matrices.  


\subsection{U-Net}
\label{sec:unetdesc}
The U-Net \cite{Ronneberger2015} is a CNN characterised by a contracting network comprising successive convolution, ReLU activation and max-pooling layers that generates low resolution feature maps, combined with an expanding network where max-pooling is replaced by upsampling to increase the resolution of the output.  Skip connections allow high resolution features generated along the contracting network to be combined with the upsampling output, thereby improving the precision of the segmentation.  We summarise the U-Net architecture using the following simplified notation
\begin{align}
    \label{eq:convunet}
    \tilde{x} &= f_U(x; \psi_u); \quad \tilde{x} \in \mathbb{R}^{d \times d \times 64}  \\
    \label{eq:logitsunet}
     \eta &= f_{1\times 1}(\tilde{x}, \psi_1); \quad \eta \in \mathbb{R}^{d \times d \times K} 
\end{align}
where the input $x \in \mathbb{R}^{d \times d}$ represents a 2D matrix of pixel values (but can also accommodate 3D images); $f_U$ denotes the combination of all $3 \times 3$ convolution, ReLU, pooling and up-convolution layers in the U-Net; $\tilde{x} \in \mathbb{R}^{d \times d \times 64}$ is a 64-channel feature map of the same resolution as the input image; $f_{1 \times 1}$ is a $1 \times 1$ convolution layer that reduces the dimensionality of the feature space to that of the desired number of classes, $K$; $\eta \in \mathbb{R}^{d \times d \times K}$ is a $K$-channel feature map of the same resolution as the original image; and $\psi = [\psi_u, \psi_1]$ is a vector of trainable parameters (e.g., convolution kernel weights).  The categorical distribution of $c_{ij}$, the class for the $ij$th pixel, is then obtained by converting the feature map $\eta$ to a probability for each channel or class using the soft-max operator, i.e.    
\begin{align}
        \label{eq:softmax}
         p_{\eta} \left(c_{ij} = k \right) = \frac{\exp \left(\eta_{ijk} \right)}{\sum_{k=1}^K \exp \left(\eta_{ijk} \right)}
\end{align}
Since we are primarily interested in binary segmentation ($K=2$), we consider a special case of the soft-max operator, the sigmoid function, to compute class probabilities,  
\begin{align}
    \label{eq:sigmoid1}
    p_\eta(c_{ij} = 1) &= \frac{1}{1 + \exp(-\beta_{ij})}, \\
    \label{eq:sigmoid2}
    p_\eta(c_{ij} = 0)  &= 1 - p_\eta(c_{ij} = 1)  = \frac{\exp(-\beta_{ij})}{1 + \exp(-\beta_{ij})}. 
\end{align}
where $\beta_{i,j} := \eta_{ij1} - \eta_{ij0}$. 
Finally, the predicted segmentation $s$, is given by a $d \times d$ integer-valued matrix of class labels from the set $\{1, 2, \cdots, K\}$, where $s_{ij}$ is typically obtained by selecting the label corresponding to the maximum class probability, 
\begin{align}
    \label{eq:segunet} 
    s_{ij} &=  \arg \max_k p_k \left(\eta_{ij} \right), \quad \text{for all} \enskip i,j.
\end{align}
This approach is known to produce unrealistic and noisy segmentations 
\cite{singh2023improving}, hence we adopt a commonly used image thresholding strategy suitable for binary segmentation whereby
\begin{align}
\label{eq:binseg}
    s_{ij} = \begin{cases}
  1  &  p_\eta(c_{ij} = 1)  \geq \tau \\
  0 & \text{otherwise},
\end{cases}
\end{align}
where $\tau$ is decision threshold.  This threshold is determined using   Otsu's method \cite{otsu1979threshold} which gives the optimal value of $\tau$ that minimises intra-class intensity variance.  Note that $\tau = 0.5$ corresponds to (\ref{eq:segunet}).

\subsection{Probabilistic U-Net}
\label{sec:probunet}
The U-Net in principle provides probabilistic outputs through independent pixel-wise categorical distributions (see \ref{eq:softmax}), although it provides a poor representation of the dependence between pixels. 
The Probabilistic U-Net \cite{kohl2018} has emerged as an empirically competitive approach to quantifying aleatoric uncertainty by making use of ideas from conditional variational auto-encoders (cVAEs) and $\beta$-VAEs \cite{Higgins2017, Kingma2019}.  This is done by incorporating a random latent factor $z \in \mathbb{R}^d$, that represents a low dimensional embedding of the image-segmentation pair, which is then used to efficiently generate samples of plausible segmentations $s$.  The distribution of $z$ is captured through the intractable posterior $p_\theta(z|{\bf x, y})$, whose form is learned from the data.  %
In cVAEs, the goal is to determine $\theta, \phi$ such that a more tractable form $q_\phi(z|{\bf x, y})$ best approximates the intractable posterior $p_\theta(z|{\bf x}, {\bf y})$ and $p_\theta(x|{\bf y})$ approximates the true conditional data distribution $p(x|{\bf y})$.  As is common in many applications of variational inference, the posterior $p_\theta(z|{\bf x, y})$ is assumed to be well approximated by a Gaussian distribution, 
\begin{align}
    q_\phi(z|{\bf x}, {\bf y}) = N(\mu({\bf x}, {\bf y}), \Sigma({\bf x}, {\bf y})),
\end{align}
where $\mu({\bf x}, {\bf y}) \in \mathbb{R}^d$, $\Sigma({\bf x}, {\bf y})$ a $d \times d$ matrix with zeros on the off-diagonal and diagonal given by $\sigma({\bf x}, {\bf y}) \in \mathbb{R}^d$.  The mean and covariance $\mu$ and $\sigma$ respectively are given by,  
\begin{equation}
\begin{aligned}
& \mu(x,y)=N N_{p o s t, \mu}\left(x,y ; \phi_\mu\right), \\
& \sigma(x,y)=N N_{p o s t, \sigma}\left(x,y ; \phi_\sigma\right),
\end{aligned}
\end{equation}
where $NN_{post, \mu}(x,y; \phi_\mu)$ indicates a neural network that takes as input an image $x$ and its corresponding segmentation $y$ and is parameterised by $\phi_\mu$; and $\phi = [\phi_\mu, \phi_\sigma]$.  The parameters $\theta, \phi$ are determined by minimising a modified version of the per-datapoint negative evidence lower bound (ELBO), given by    
\begin{align}
    \label{eq:objfunc}
    \underbrace{-\mathbb{E}_{q_\phi(z| x,  y)} [\log p_\theta (y| x, z)]}_{\text{reconstruction term}} + \beta \underbrace{KL(q_\phi( z| x, y) || p_\theta( z| x))}_{\text{regularisation term}},
\end{align}
where $\beta$ is a tuning parameter controlling the influence of the regularisation term and $p_\theta({\bf z}|{\bf x})$ is the conditional prior on $z$ which is also learnt from the data.  It takes a similar form to the approximate posterior except that only images ${\bf x}$ are taken as inputs, i.e. $p_\theta(z|{\bf x}) = N(\mu_{prior}({\bf x}), \Sigma_{prior}({\bf x}))$, with 
\begin{equation}
\begin{aligned}
& \mu_{\text {prior }}(x)=N N_{\text {prior} , \mu}\left(x ; \theta_\mu\right) \\
& \sigma_{\text {prior }}(x)=N N_{\text {prior }, \sigma}\left(x ; \theta_\sigma\right),
\end{aligned}
\end{equation}
and $\theta = [\theta_\mu, \theta_\sigma]$ denotes the weights and biases of the prior neural networks.  The reconstruction term in \ref{eq:objfunc} is equivalent to the expected cross-entropy, which penalises differences between the ground truth segmentation $y$ and generated segmentation $s$. In summary, the entire set of parameters $\{\theta, \psi, \phi\}$ is jointly learned by minimising an empirical approximation of the loss function \ref{eq:objfunc} assuming independent and identically distributed (iid) data, given by 
\begin{align}
    \label{eq:probunetloss}
   \sum_{n=1}^N \left( \frac{1}{M} \sum_{m=1}^M -\log p_{\eta^{m,n}} (c = y^n) \right) + \beta KL(q_\phi(z|x^n, y^n) || p_\theta(z|x^n))
\end{align}
where $p_{\eta^{m,n}} (c = y^n) = \prod_{ij} p_{\eta^{m,n}} (c_{ij} = y_{ij}^n)$ and $p_{\eta^{m,n}} (c_{ij} = y^n_{ij})$ is the class probability of the $i,j$th pixel corresponding to the $n$th true segmentation $y^n$, computed using the softmax or sigmoid operator (\ref{eq:softmax} or \ref{eq:sigmoid1}-\ref{eq:sigmoid2} respectively). 
The feature map  corresponding to the $n$th image is obtained using a minor modification to the standard U-Net, 
\begin{align}
    \label{eq:u-netsing}
    \tilde{x}^n &= f_U(x^n; \psi_u); \\ 
    \label{eq:zsamp}
     z^{m,n} &\sim q_\phi(z|x^n, y^n);  \\
     \label{eq:activprobsing}
     \eta^{m,n} &= [f_{1\times 1}(\tilde{x}^n, \psi_1); E_{d \times d}(z^{m,n})] 
\end{align}
where $E_{d \times d}$ denotes the expansion operation that expands $z^n$ by repeating its values to match the dimensions of $f_{1\times 1}(\tilde{x}^n, \psi_1)$.  
Further details on training are given in Section \ref{sec:traineval}.  Once trained, a set of $M$ plausible segmentations for a new image $\mathcal{X}$ can be obtained by evaluating \ref{eq:u-netsing}-\ref{eq:activprobsing} $M$ times with \ref{eq:zsamp} replaced by 
\begin{align}
    z^i &\sim p_\theta(z|\mathcal{X})
\end{align}
The resulting feature map for each sample is then converted to a segmentation using the same procedure as detailed in section \ref{sec:unetdesc}. 
An overview of the method can be found in Figure \ref{fig:baselines}b.

\subsection{Diversified and Personalised Multi-rater Segmentation (D-Persona)}

D-Persona \cite{wu2024dpersona} was proposed to address the challenges of multi-rater medical image segmentation, where annotation ambiguity arises due to data uncertainties and diverse observer preferences. Unlike traditional methods, D-Persona aims to generate both diversified and personalised segmentation results by employing a novel two-stage framework.

\paragraph{Stage I: Diversified Segmentation}
The first stage learns a common latent space to capture the variability across multiple expert annotations. Using a Probabilistic U-Net as the base model, this stage introduces a bound-constrained loss to enhance segmentation diversity. 

The latent space is defined by prior and posterior distributions:
\begin{equation}
\begin{aligned}
\mu_{\text{prior}}, \sigma_{\text{prior}} &= F_{\text{prior}}(X), \\
\mu_{\text{post}}, \sigma_{\text{post}} &= F_{\text{post}}(X, \mathcal{A}_{\text{set}}),
\end{aligned}
\end{equation}

where \( F_{\text{prior}} \) and \( F_{\text{post}} \) are neural encoders, \( X \) represents the input, and \( \mathcal{A}_{\text{set}} \) denotes the set of expert annotations. The Kullback-Leibler divergence aligns these distributions:
\begin{align}
L_{\text{KL}} = \text{KL}(D_{\text{prior}}(X) \| D_{\text{post}}(X, \mathcal{A}_{\text{set}})).
\end{align}

To encourage diversity, intersections and unions of annotations (\(\mathcal{A}_{\text{bound}} = \{\mathcal{A}_{\text{inter}}, \mathcal{A}_{\text{union}}\}\)) are computed, and a bound loss is applied:
\begin{align}
L_{\text{bound}} = \text{DSC}(P_{\text{prior, inter}}, \mathcal{A}_{\text{inter}}) + \text{DSC}(P_{\text{prior, union}}, \mathcal{A}_{\text{union}}),
\end{align}
where \( P_{\text{prior, inter}} \) and \( P_{\text{prior, union}} \) are predictions derived from the prior distribution. The overall loss for Stage I is:
\begin{align}
L_{\text{stage 1}} = L_{\text{KL}} + \alpha L_{\text{seg}} + \beta L_{\text{bound}},
\end{align}
where \( L_{\text{seg}} \) is a segmentation loss based on randomly sampled latent codes.

\paragraph{Stage II: Personalised Segmentation}
In the second stage, D-Persona generates personalised segmentations by learning expert-specific prompts from the common latent space. For each expert \( i \), a projection head \( F_{\text{proj}, i} \) maps the U-Net features into expert-specific latent codes:
\begin{align}
z_{\text{exp}, i} = \text{Pooling}[F_{\text{proj}, i}(F_{\text{backbone}}(X))].
\end{align}
To ensure these latent codes remain within the learned latent space, a cross-attention mechanism aligns \( z_{\text{exp}, i} \) with a prior bank:
\begin{align}
\tilde{z}_{\text{exp}, i} = \text{Softmax}(z_{\text{exp}, i}^\top \cdot z_{\text{prior bank}}) \cdot z_{\text{prior bank}}^\top.
\end{align}

The final segmentation for expert \( i \) is obtained by:
\begin{align}
P_{\text{exp}, i} = F_{\text{head}}(\tilde{z}_{\text{exp}, i}, F_{\text{backbone}}(X)),
\end{align}
and supervised using:
\begin{align}
L_{\text{corr, seg}} = \sum_{i=1}^n \text{DSC}(P_{\text{exp}, i}, A_i),
\end{align}
where \( A_i \) is the annotation corresponding to expert \( i \).

\subsection{Collectively Intelligent Medical Diffusion (CIMD)}

Collectively Intelligent Medical Diffusion (CIMD) \cite{rahman2023cimd} was proposed to address the ambiguity inherent in medical image segmentation by utilising the stochastic nature of diffusion models. Unlike the Probabilistic U-Net and its hierarchical variants, CIMD generates multiple plausible segmentation masks without the need for an additional network to encode prior distributions. The method directly models ambiguity during training using a novel hierarchical diffusion framework.

The diffusion process consists of two phases: a forward process that iteratively adds Gaussian noise to the segmentation mask \( x_0 \), and a reverse process that denoises the perturbed mask to recover clean annotations:
\begin{equation}
\begin{aligned}
\label{eq:CIMD_forward_reverse}
    q(x_t|x_{t-1}) &= \mathcal{N}(x_t; \sqrt{\alpha_t} x_{t-1}, (1-\alpha_t)I), \\
    p_\theta(x_{t-1}|x_t) &= \mathcal{N}(x_{t-1}; \mu_\theta(x_t, t), \Sigma_\theta(x_t, t))
\end{aligned}
\end{equation}
where \( \alpha_t \) is a noise scheduling parameter, and \( \mu_\theta \) and \( \Sigma_\theta \) are learned by the model.

To capture the ambiguity of medical segmentations, two neural networks are used:
\begin{enumerate}
\item The Ambiguity Modelling Network (AMN), which models the distribution of ground truth annotations as:
\begin{align}
Q(z|b, x_b) = \mathcal{N}(\mu(b, x_b; \nu), \sigma(b, x_b; \nu)),
\end{align}
where \( \nu \) represents trainable parameters.
\item The Ambiguity Controlling Network (ACN), which models the ambiguity of predictions during the reverse process as:
\begin{align}
P(z|b, \hat{x}_b) = \mathcal{N}(\mu(b, \hat{x}_b; \omega), \sigma(b, \hat{x}_b; \omega)),
\end{align}
where \( \omega \) denotes trainable parameters.
\end{enumerate}

The loss function combines components to optimise the reverse process and align the ambiguity distributions of AMN and ACN:
\begin{align}
\label{eq:cimd_loss}
L_\text{total} = L_\text{simple} + \lambda L_\text{vlb} + \beta D_\text{KL}(Q(z|b, x_b) \| P(z|b, \hat{x}_b)),
\end{align}
where \( L_\text{simple} \) minimises the noise prediction error, \( L_\text{vlb} \) is a variational lower bound, and the KL divergence term \( D_\text{KL} \) penalises the mismatch between AMN and ACN distributions.

Unlike traditional models, CIMD introduces stochasticity at each hierarchical level, enabling the generation of diverse and accurate segmentation masks. These masks align with ground truth distributions and inherently capture inter-expert variability without requiring prior networks. The sampling process during inference iteratively refines segmentation:
\begin{align}
x_{t-1} \leftarrow \frac{1}{\sqrt{\alpha_t}} \left(x_t - \frac{1-\alpha_t}{\sqrt{1-\gamma_t}} f_\theta(x_t, t) \right) + \gamma_t z,
\end{align}
where \( z \sim \mathcal{N}(0, I) \) and \( t \in [1, T] \).

This framework significantly improves segmentation diversity and accuracy, particularly in medical contexts requiring collective intelligence for diagnostics.

\subsection{Multi-head Variational Inference U-Net (MH VI U-Net)}
The MH-VI U-Net \cite{Fuchs2022} is a variational inference based extension of the multi-headed U-Net
which aims to improve out-of-distribution robustness.  The multi-head approach consists of a base model (e.g., a U-Net) and an ensemble of ``heads'', typically a series of convolutions in the final layers of the U-Net.  The MH-VI U-Net aims to improve on the classical MH U-Net by replacing regular convolutions with Bayesian convolutions in the heads. 
The classical MH U-Net allows only for maximum likelihood estimation of the weights at each head.  \\ 
\\
 The base model is taken to be a reduced parameter version of the U-Net, and the depth of each head is typically restricted by the number of heads due to memory constraints. Variational inference is then employed to efficiently approximate the posterior distribution of the weights in head $h$, $w_h$ which is approximated by $q_{\theta_h}(w_h)$.  Here $\Theta = [\theta_1, \theta_2, \cdots, \theta_H]$ denotes the variational parameters to be optimised, obtained by minimising a linear combination of the per head negative ELBO and the joint ELBO over all $H$ heads given by
 \begin{equation}     
    \lambda \mathcal{L}_{\Theta}({\bf x,y}) + (1 - \lambda) \frac{1}{H}\sum_{h=1}^H \mathcal{L}_{\theta_h}({\bf x,y})
 \end{equation}
where $\mathcal{L}_\mu({\bf x,y})$ is the negative ELBO, i.e.
\begin{align}
    \mathcal{L}_\mu({\bf x,y}) := - \mathbb{E}_{q_\mu(w)} [\log p({\bf y}| {\bf x}, w)] + \beta KL(q_\mu(w) || p(w)) 
\end{align}
where $\beta$ is the temperature, $w$ is a vector of the weights of the U-Net and $\mu$ denotes the variational parameters to be optimised.  The tuning parameter $\lambda$ can be seen as a bridge between classical deep ensemble training $(\lambda = 0)$ and joint estimation $(\lambda = 1)$.  While both the VI-MH U-Net and the Probabilistic U-Net rely on variational inference, they primarily differ in terms of how randomness is introduced into the U-Net: in the VI-MH U-Net this is via the weights of the network, compared to in the probabilistic U-Net where this is done on some low dimensional latent feature map.

\subsection{Stochastic Segmentation Networks (SSN)}
Stochastic Segmentation Networks \cite{Monteiro2020} were proposed to model aleatoric uncertainty primarily due to integrader variability.  Unlike the Probabilistic U-Net, the SSN can be more easily used with any neural network based segmentation model.  The logits (i.e. inputs to the softmax) $\eta \in \mathbb{R}^{d^2.K \times 1}$, represented here as a flattened vector, are modelled as a Gaussian random variable whose mean and covariance are modelled by neural networks, 
\begin{align}
\label{eq:p_eta_x}
    p_\gamma(\eta|x) = N(\mu(x), \Sigma(x)),
\end{align}
where 
\begin{align}
    \Sigma(x) = PP^T + D,
\end{align}
and $P \in \mathbb{R}^{d^2.K \times R}$ is known as a covariance factor and $R$ controls the rank of the matrix and $D \in \mathbb{R}^{d^2K \times d^2K}$ is a diagonal matrix with $\sigma$ on the diagonal.  The mean, covariance factor and diagonal are represented by neural networks,
\begin{equation}
\begin{aligned}
    \mu(x) &= NN_\mu(x; \gamma_\mu) \\
    \sigma(x) &= NN_\sigma(x; \gamma_\sigma) \\
    P(x) &= NN_P(x; \gamma_P).
\end{aligned}
\end{equation}
where the parameters $\gamma = [\gamma_\mu, \gamma_\sigma, \gamma_P]$ are learnt during training.  Note that unlike in the Probabilistic U-Net, the uncertainty on the logits is not updated according to the segmentation masks $y$ in the training data, in other words, there is no Bayesian updating of a prior to a posterior. The parameters of the segmentation network, $\psi$, and $\gamma$ are jointly learned by minimising the negative log-likelihood over the entire dataset, assuming iid data, 
\begin{align}
    \label{eq:SSNobj}
    -\log \left( p({\bf y} | {\bf x}) \right) &= -\sum_{n=1}^n \log \left( \mathbb{E}_{p_\gamma(\eta^n| x^n)} [p( y^n|\eta^n)] \right) \\
    &\approx -\sum_{n=1}^n \log \left( \frac{1}{M} \sum_{m=1}^M \prod_{i,j} p_{\eta^{n,m}} (c_{ij} = y^n_{ij}) \right) 
\end{align}
where $\eta^{n,m} \sim N(\mu( x^n), \Sigma(x^n))$ as defined in \ref{eq:p_eta_x}. 
 Note that \ref{eq:SSNobj} takes a similar form to \ref{eq:probunetloss} but without a regularisation term due to the lack of the Bayesian update.  This may potentially lead to a less tractable optimisation problem that may have difficulty converging \cite{poggio2020complexity}. 



\subsection{Monte Carlo Dropout}
Monte Carlo Dropout (MCDO) is a widespread tool for approximating model uncertainty in neural networks by randomly removing layers during both training and prediction phases \cite{kendall2015bayesian,gal2016dropout}. 
In the context of the U-Net, MCDO can be implemented by adding spatial dropout layers \cite{tompson2015efficient} after every encoding (pair of convolution layers with batch normalisation and ReLU activations, followed  by average pooling) and decoding block (transposed convolution, followed by a pair of convolution layers wit batch normalisation and ReLU).  The key idea behind MCDO is to randomly switch on and off neurons (i.e. set their weights to zero) according to a pre-specified probability.  Running this procedure $M$ times allows one to generate $M$ segmentations. 
However, it is noteworthy  that even though utilising the Dropout function is computationally inexpensive, there are a number of intrinsic shortcomings of MCDO, such as unreliable uncertainty quantification, subpar calibration, and inadequate out-of-distribution detection \cite{Fuchs2022}.

\section{PULASki - Probabilistic Unet Loss Assessed through Statistical distances} 
\label{sec:pulaski}

\begin{figure*}[htbp!]
    \centering
  \includegraphics[width=13cm,height=9cm]{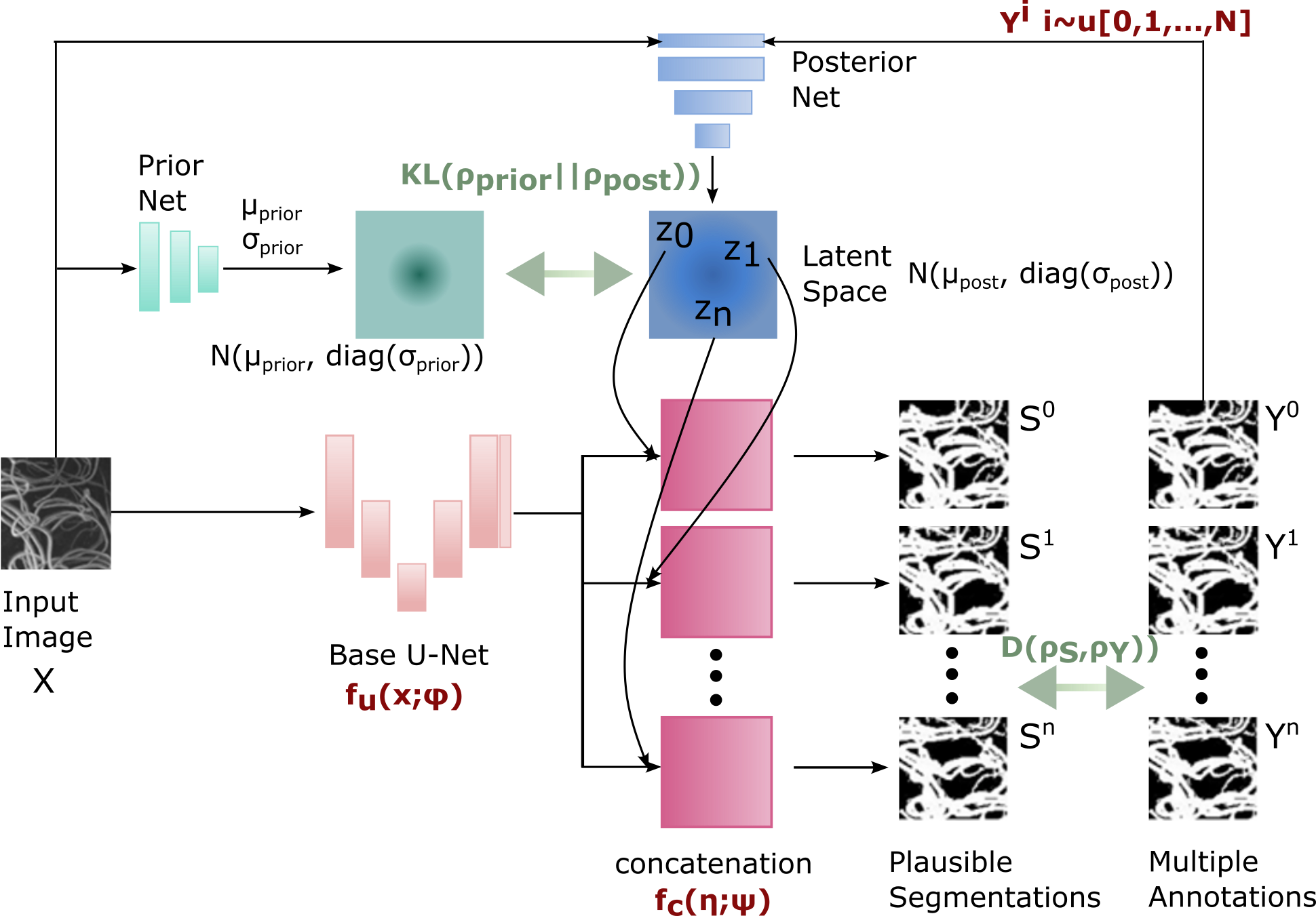}
  \caption{Schematic of the proposed PULASki method.}
  \label{fig:proposedmethod}
\end{figure*}

Our main focus is to improve the capability of U-Net based image segmentation methods to capture predicted aleatoric uncertainty, primarily due to inter-rater variability. Inter-rater variability is a persistent challenge in biomedical segmentation, often driven by factors such as limited image clarity (e.g., lower signal-to-noise ratio, presence of artefacts) and the inherent complexity of the structures being segmented (e.g., small cerebral vessels).  In the original formulation of the Probabilistic U-Net, the reconstruction term in the ELBO \ref{eq:objfunc} can be seen as an empirical approximation of the expected cross-entropy,
\begin{align}
    \mathbb{E}_{z \sim q_\phi(z|x,y)}[H(p(y|x), p_\theta(y|z,x)] 
\end{align}
where 
\begin{align}
    H(p(y|x), p_\theta(y|z,x)) = -\sum_{y \in \mathcal{Y}} p(y|x)  \log p_\theta(y|z, x)  
\end{align}
and $\mathcal{Y}$ is the set of all possible segmentations for a given image. The standard cross-entropy has several known shortcomings for class-imbalanced datasets and for multi-label classification problems \cite{abraham2019novel,rezaei2020addressing,tian2022striking}.  An empirical approximation of the cross-entropy between $p(y|x)$ and $p_\theta(y|z,x)$ when used to approximate the ELBO is indistinguishable from the standard ELBO (\ref{eq:probunetloss}) when multiple labels are treated as conditionally independent data.  
This makes it poorly suited for learning the conditional distribution $p(y|x)$.   

We propose to modify the loss function \ref{eq:probunetloss} in the probabilistic U-Net by replacing the expected cross-entropy term with a more general statistical distance between $p_\theta(y|x)$ and $p(y|x)$.  Various choices are possible, and we focus our attention on those that have the potential to more efficiently capture features of $p(y|x)$ when only samples from it are available.  Our proposed loss function (per data point) is 
\begin{align}
   \mathbb{E}_{ z \sim q_\phi(z| x,y)} \left[ D(p( y| x), p_\theta(y|x,z)) \right] +  \beta KL(q_\phi( z| x, y) || p_\theta( z| x))
\end{align}
where $D(p,q)$ is an appropriate statistical distance between two probability distributions $p,q$.  In practice, distances are usually approximated using the empirical distributions; Samples from $p_\theta(y|x,z)$ can be easily generated during training by applying the same latent vector operations as in the Prob U-net $M$ times (see \ref{eq:u-netsing} - \ref{eq:activprobsing}).  An empirical approximation of $p(y|x)$ is available whenever there are multiple plausible segmentations per image.  A schematic of the method can be found in Figure \ref{fig:proposedmethod}. 
 Our proposed method bears some resemblance to wasserstein autoencoders \cite{Tolstikhin2018}, although our primary focus is on varying the reconstruction loss for conditioned generation whereas their focus is on alternative regularisation terms (i.e. the KL term). 

\begin{algorithm}[tb]
\caption{PULASki Training}
\begin{algorithmic}
\Require Training dataset $\{(x_i, \{y_i^j\}_{j=1}^M)\}_{i=1}^N$ with $N$ images and $M$ ground-truth segmentations per image
\Require Statistical distance function $D(P, Q)$
\Require Weight for KL divergence term, $\beta$
\Ensure Trained model parameters

\For{each training iteration}
    \State Sample a minibatch of input images $\{x_b\}_{b=1}^B$ and corresponding ground-truth segmentations $\{\{y_b^j\}_{j=1}^M\}_{b=1}^B$
    \For{each training iteration}
    \State Sample a minibatch of input images $\{x_b\}_{b=1}^B$ and corresponding ground-truth segmentations $\{\{y_b^j\}_{j=1}^M\}_{b=1}^B$
    \For{each image $x_b$ in the minibatch}
        \State Encode $x_b$ to obtain prior parameters: $p_\theta(z|x_b)$
        \State Encode $x_b$ and each $y_b^j$ to obtain posterior parameters: $q_\phi(z|x_b, y_b^j)$
        \State Sample $M$ latent variables $\{z_b^j\}_{j=1}^M$ from the posterior distributions $q_\phi(z|x_b, y_b^j)$
        \State Generate $M$ predicted segmentations $\{\hat{y}_b^j\}_{j=1}^M$ using the decoder: $\hat{y}_b^j = \text{Decoder}(x_b, z_b^j)$
        \State Calculate statistical distance loss: $L_{\text{StatDist}} = D(\{\hat{y}_b^j\}_{j=1}^M, \{y_b^j\}_{j=1}^M)$
        \State Calculate KL divergence for each pair: $L_{\text{KL}} = \frac{1}{M} \sum_{j=1}^M \text{KL}(q_\phi(z|x_b, y_b^j) \parallel p_\theta(z|x_b))$
        \State Compute total loss for $x_b$: $L_b = L_{\text{StatDist}} + \beta \cdot L_{\text{KL}}$
    \EndFor
    \State Compute average loss over the minibatch: $L = \frac{1}{B} \sum_{b=1}^B L_b$
    \State Update model parameters to minimise $L$ using backpropagation
\EndFor
    \State Compute average loss over the minibatch: $L = \frac{1}{B} \sum_{b=1}^B L_b$
    \State Update model parameters to minimise $L$ using backpropagation
\EndFor
\end{algorithmic}
\end{algorithm}

Crucially, all statistical distances considered here are differentiable, so that they can be used as loss-functions in standard gradient-based optimisation methods such as ADAM.  
Furthermore, distances focusing on both distributional quality as well as local geometric features are chosen.  Broadly speaking, FID and Sinkhorn divergence are used to measure global distributional differences between samples of images, whereas the Hausdorff divergence is primarily used to measure differences between sets of points, thereby emphasising differences in local geometric structure. Therefore, both FID and Sinkhorn are more widely used in the image generation literature (e.g. GANs and diffusion modelling) whereas Hausdorff divergence is used in used in shape analysis and object detection in the medical imaging context. 
 The Hausdorff divergence is much more sensitive to differences in local image features such as edges and contours, making it a useful tool for comparing segmentations.  The considered distances (FID, Hausdorff and Sinkhorn) are described in further detail below.


\subsection{Frechet Inception Distance (FID)}
The FID \cite{Heusel2017} is a pseudometric based on the Wasserstein-2 distance and is widely used particularly in training GANs for image generation.  Computing the Wasserstein-2 distance between two arbitrary empirical distributions can be computationally prohibitive in DL applications which require repeated evaluations during training. Calculation of the FID requires approximating the distributions by a Gaussian, which then allows one to exploit the existence of a closed form solution to the Wasserstein-2 distance for Gaussian distributions, making it fast to implement.  It provides a reasonable approximation for distributions whose first two moments exist, although it can produce biased estimates \cite{Binkowski2018}. 



In the image generation or segmentation context, the FID is computed as 
\begin{align}
    D_{FID} := |\mu - \hat{\mu}|_2^2 + Tr(\Sigma + \hat{\Sigma} - 2(\Sigma^{1/2} \hat{\Sigma} \Sigma^{1/2})^{1/2})
\end{align}
where $\hat{\mu} \in \mathbb{R}^{v}$ and $\hat{\Sigma} \in \mathbb{R}^{v \times v}$ are the sample mean and covariance respectively of $\{ a^m\}_{m=1:M}$ where $a^m$ denotes a flattened (i.e. vectorised) version of the matrix $\eta^m$ as defined in \ref{eq:activprobsing} of length $v = d\times d \times K$. 
Likewise, $\mu$ and $\Sigma$ denote the sample mean and covariance respectively of the final activations (i.e. logits) from an Inception Net V3 \cite{szegedy2016rethinking} which takes the plausible segmentations rather than raw images as inputs. FID, employing Inception Net V3 trained on ImageNet, provides a viable metric for medical images by quantifying the statistical alignment of low-level visual features (e.g., edges, textures) that are inherently universal across image domains. Since FID evaluates the distributional similarity of these domain-agnostic patterns, rather than relying on semantic or medical specificity, it offers a pragmatic solution for assessing generative models in medical imaging contexts. Since the inception net is only trained on 2D images, we only make use of the FID for the experiments involving 2D images.  



\subsection{Hausdorff Divergence}  
The Hausdorff \textit{distance} is a way of measuring distances between subsets of metric spaces (rather than measure spaces).  It has traditionally been used in computer vision to measure the degree of similarity between two objects, such as image segmentations, by measuring distances between points of the two sets.  The traditional Hausdorff distance can be lifted to measuring statistical distance between probability measures.  In the case of discrete measures $\vartheta, \xi$ on $\mathbb{R}^d$ given as weighted point clouds i.e. 
\begin{align}
    \label{eq:discmeas}
    \vartheta = \sum_{i=1}^N \vartheta_i \delta_{x_i}; \quad \xi = \sum_{i=1}^N \xi_i \delta_{y_i};
\end{align}
where $\delta_{x}$ is the Dirac delta function centred at $x$, the Hausdorff \textit{divergence} is defined (as per definition A.2 in \cite{Feydy2020}) as
 \begin{align}
     d_{H}(\vartheta, \xi) := \frac{1}{2} \langle \vartheta - \xi, \nabla F_\epsilon(\vartheta) - \nabla F_\epsilon (\xi) \rangle.  
 \end{align}
The Hausdorff divergence is equivalent to the symmetric Bregman divergence induced by $F_\epsilon$, a strictly convex functional, where $F_\epsilon$ is the Sinkhorn negentropy, defined as:
 \begin{equation}
    F_{\epsilon}(\vartheta) := -\frac{1}{2} \mathrm{OT}_{\epsilon}(\vartheta, \vartheta)
\end{equation}
where $OT_\epsilon$ is the entropy-regularised optimal transport cost, given by: 
\begin{equation}
\label{eq:ote}
OT_\epsilon(\vartheta, \xi) = \min_{\pi \in M^+(X \times X)} \left\langle \pi, C(x, y) \right\rangle + \epsilon KL(\pi || \vartheta \otimes \xi)
\end{equation}
and we let $C(x, y) = \frac{1}{2}\|x-y\|_2^2$,i.e. the squared Euclidean distance (corresponding to the Wasserstein-2 distance).  
 
 The above is more specifically referred to as the $\epsilon$-Hausdorff divergence, as it is based on $OT_\epsilon$ which only produces an $\epsilon$-approximate solution to the exact OT problem.  The $\epsilon$-Hausdorff divergence is generally cheaper to compute than the closely related $\epsilon$-Sinkhorn divergence or debiased Sinkhorn divergence (see Section \ref{subsect:de-biasSink}) as it relies on solving the symmetric OT problem $OT_\epsilon(\vartheta, \vartheta)$ rather than $OT_\epsilon(\vartheta, \xi)$. 


 
\subsection{de-biased Sinkhorn divergence}
\label{subsect:de-biasSink}
A more accurate but computationally expensive approximation of the $p$-Wasserstein distance is the so-called de-biased Sinkhorn divergence \cite{Ramdas2017} 
\begin{align}
    \label{eq:sinkhorn}
    S_\epsilon(\vartheta, \xi) := \text{OT}_\epsilon(\vartheta, \xi) - \frac{1}{2} \text{OT}_\epsilon(\vartheta, \vartheta) - \frac{1}{2} \text{OT}_\epsilon(\xi, \xi) 
\end{align}
where $\vartheta, \xi$ are any positive measures and $\text{OT}_\epsilon$ is the entropic-regularised form of the optimal transport loss \ref{eq:ote}. 
This divergence has been used in training generative models (See e.g. \cite{genevay18a} ).
We utilise the method proposed in \cite{Feydy2020} which makes use of a multiscale Sinkhorn algorithm to compute $S_\epsilon(\vartheta, \xi)$.  

\section{Experiments}
\label{sec:experiments}

\subsection{Datasets and Labels}

\begin{figure*}
\centering
\includegraphics[width=0.85\textwidth]{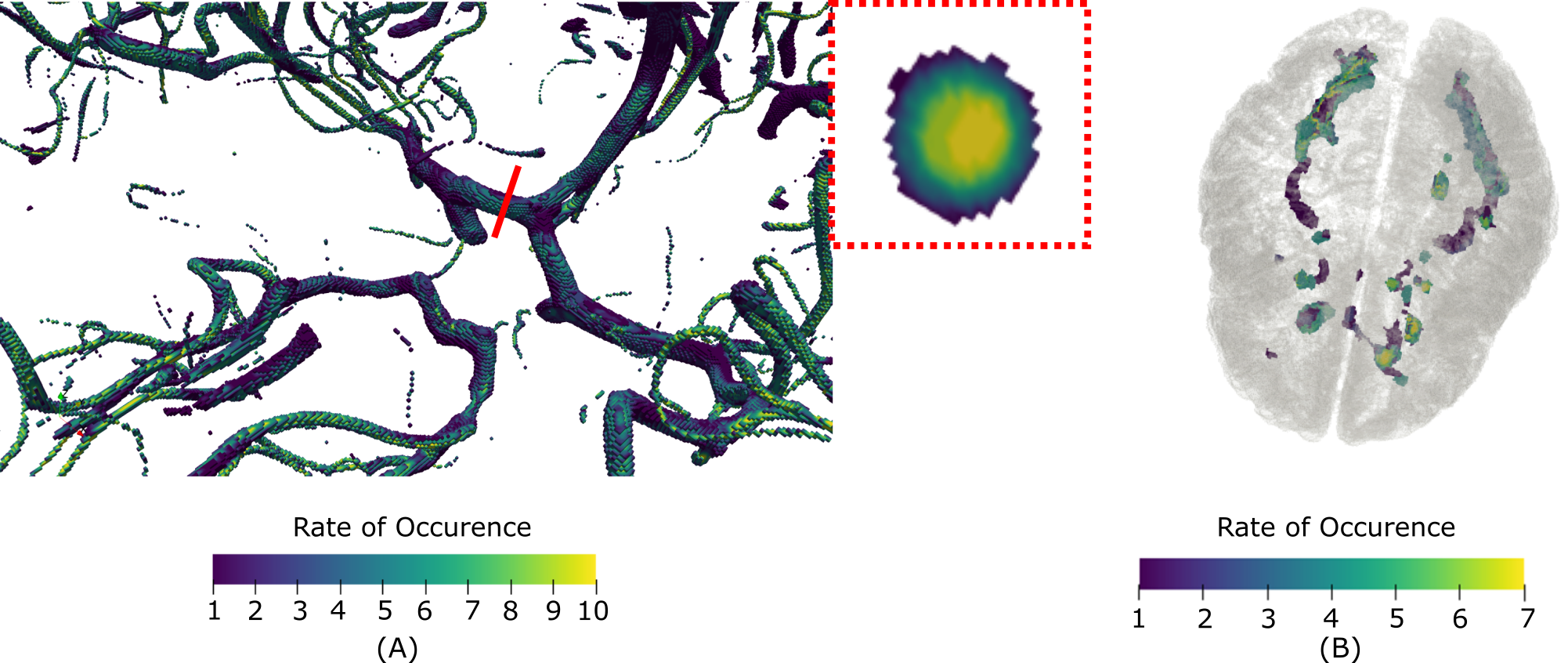}
  \caption{Rate of occurrence (RoO) of labels across multiple annotations. The brighter (yellow) the voxel appears, the more often it
is labelled as vessel (A) or lesion (B) across the available annotations.   (A): Ten segmentations per image from one subject in OpenNeuro’s ‘StudyForrest’ dataset were generated using the Frangi filter.  
A vessel cross section at the location of the red line is displayed within the red square. 
(B): Multiple Sclerosis lesion segmentations from 7 expert annotators for one subject.}
    \label{Fig:RoO_OD}
\end{figure*}

To be able to validate the proposed approach, the OpenNeuro’s ‘StudyForrest’ dataset which contains 20 subjects was used for vessel segmentation. It contains 7T MRA-ToF full brain volumetric images with [480 × 640 × 163] matrix size in NIFTI format recorded with a Siemens MR scanner at 7-Tesla using a 3D Multiple Overlapping Thin Slab Acquisition Time of Flight angiography sequence (0.3 mm isotropic voxel size) \cite{hausler2021studyforrest}. Training the network requires multiple plausible segmentations. For the ‘StudyForrest’ dataset, multiple manual labels did not exist. Therefore, ten synthetic plausible segmentations per case have been generated using the Frangi filter \cite{frangi1998}. The Frangi scale range has been set to $\sigma$ $\sim$ [1 8] and a step size between the sigmas of 2 was chosen. Following parameters, known as vesselness constants, have been sampled from truncated normal distributions: (i) $\mathrm{Frangi}_{\alpha}$ $\sim$ TN(0.5 1) determines if its a line (vessel) or plane like structure; (ii) $\mathrm{Frangi}_{\beta}$ $\sim$ TN(0.5 1) determines the deviation from a blob like structure; (iii) $\mathrm{Frangi}_{C}$ $\sim$ TN(500 100) gives a threshold between eigenvalues of noise and vessel structures. The resulting "inter-rater" agreement of the generated dataset was calculated using $K\alpha$ (discussed later in Sec. \ref{sec:traineval}) and has a value of 0.40. The labels generated during this research are made available publicly as part of the SMILE-UHURA dataset \cite{chatterjee2024smile}.

For further validating the proposed method, 3T FLAIR MRIs from the publicly available multiple sclerosis (MS) segmentation dataset, that was used for MICCAI 2016 challenge \cite{commowick2021multiple}, containing MRI scans of 53 patients and their corresponding manual segmentations of the lesions by seven experts, was utilised. After thorough examination of expert voting, one subject was excluded due to significant discrepancies in expert annotations. In this case, two graders determined that no voxels represented an MS lesion, four experts identified only a few voxels as MS tissue, and one grader marked nearly the entire brain. Accurate identification of multiple sclerosis lesions is difficult due to variability in lesion location, size and shape -- which is also reflected in the resulting very low inter-rater agreement. The inter-rater agreement was assessed using $K\alpha$ for the MS dataset, resulting in an agreement of 0.60. However, when considering only brain regions where at least one grader noted an MS occurrence, the alpha decreased to 0.28.

The Rate of Occurence (RoO), illustrating the frequency of a voxel being labeled as a vessel or MS lesion in the annotation is displayed in Figure \ref{Fig:RoO_OD}. The RoO serves as a visual representation of the data variability, complementing the $K\alpha$ coefficient. The overall darker shade of the CoW region in the vessel image indicates high variability in the underlying plausible segmentations. The cross-sectional view of the vessel reveals that variability increases with distance from the center of the vessel. Smaller vessels exhibit comparatively lower variability, even in the outer regions, when compared with the larger vessels within the CoW region. In the MS dataset, certain expansive regions, denoted by dark blue areas, are identified as lesions by only one grader. Voxels at the center of the MS region are consistently marked as MS by (almost) all graders.

\subsection{Implementation}
\label{sec:implementation}

There are two methods for managing 3D volumetric images: (i) processing them in 2D, treating each slice as an individual data point, or (ii) working directly in 3D. However, due to computational limitations, it is often impossible to work with the full 3D volume. Hence, patch-based approaches are adapted. These patches can be 3D patches (3D sub-blocks from the 3D volume), or 2D patches (2D slices taken a particular imaging plane). 2D models are less computationally hungry, while 3D patches provide better spatial context encoding possibility. This research explores both these directions -- working with 2D slices following the acquisition plane using 2D models utilising 2D convolution operations and working with 3D patches employing 3D models with 3D convolution operations. 

The PULASki method was implemented by extending the pipeline \footnote{DS6 Pipeline: \url{https://github.com/soumickmj/DS6}} available from DS6 \cite{Chatterjee2020}. It is based on PyTorch \cite{paszke2019pytorch} and used TorchIO \cite{perez2021torchio} for creating 3D patches or extracting 2D slices from the available 3D volumes.  All the statistical distances used in PULASKi (discussed in Sec. \ref{sec:pulaski}) were calculated using the GeomLoss\footnote{GeomLoss on GitHub: \url{https://github.com/jeanfeydy/geomloss}} \cite{feydy2019interpolating}. The weights and biases of all the models (proposed and baselines, including their sub-models -- e.g., prior net),  were initialised by sampling a uniform distribution $\mathcal{U}(-\sqrt{k}, \sqrt{k})$, where $k=\frac{\text { groups }}{C_{\text {in }} * \prod_{i=0}^1 \text { kernel\_size }[i]}$ for the convolutional layers and $k=\frac{1}{in\_features}$ for the linear layers.

In this implementation, the latent space was formulated using three convolutional channels, each with a single pixel/voxel (shape 1x1 for 2D and 1x1x1 for 3D), resulting in the size of the latent variable being three. These channels were expanded to the original input size prior to 'injecting' them into the last layer of the base U-Net of the probabilistic U-Nets (PULASki methods, as well as baseline Prob UNet). This 'injection' is performed by concatenating a sampled value from the latent space with the output of the penultimate layer of the base U-Net, and then supplying it to the final layer, which is a convolutional layer with a kernel size of 1x1 or 1x1x1 for the 2D and 3D versions, respectively.

The baseline probabilistic U-Net experiments were performed by adapting the PyTorch implementation\footnote{PyTorch implementation of Probabilistic UNet baseline: \url{https://github.com/jenspetersen/probabilistic-unet}} available from \cite{petersen2019deep}. As the loss function for the baseline probabilistic UNet, as well as for the MCDO UNet, Focal Tversky Loss (FTL) \cite{abraham2019novel} was employed. Cross-entropy (CE) loss used in the original probabilisitic UNet work was replaced with FTL as CE might not be optimal when there is a big imbalance between the positive (1) and negative (0) classes in the image mask as CE treats them both equally -- a common scenario in medical image segmentation tasks \cite{rajaraman2021novel,tian2022striking}. A balanced version of CE loss (weighted CE) might combat this issue \cite{tian2022striking}. Dice loss, on the other hand, is a commonly used loss function in medical image segmentation tasks, providing a better trade-off between precision and recall \cite{abraham2019novel}. FTL, designed specifically for medical image segmentation tasks, achieves an even better trade-off between precision and recall when training on small structures \cite{abraham2019novel}, and has also been shown to outperform CE, weighted CE, and Dice \cite{rajaraman2021novel}.

The SSN method was performed using the DeepMedic convolutional neural network \cite{KAMNITSAS201761} as backbone, as was done in the original paper \cite{Monteiro2020}. DeepMedic is a dual pathway, 11 layers deep, three-dimensional convolutional neural network, that was developed for brain lesion segmentation. The network has two parallel convolutional pathways that process the input images at multiple scales simultaneously, capturing both local and larger contextual information, and employs a novel training scheme that alleviates the class-imbalance problem and enables dense training on 3D image segments. Contrary to U-Nets, which condense and then expand the input to produce the final output, DeepMedic employs a fully convolutional network that processes the input at different resolutions through two parallel pathways, then uses a series of convolutional and pooling layers to extract features at different scales, and finally combines these features through concatenation and further convolutional layers to provide the output. 

MH-VI U-Net \cite{Fuchs2022}, was implemented by adapting the official code\footnote{MH-VI U-Net official code: \url{https://github.com/MECLabTUDA/VIMH}}. All parameters, including four heads with a $\lambda$ of $0.5$, were kept the same as the original implementation. Similarly, D-Persona \cite{wu2024dpersona} and CIMD \cite{rahman2023cimd} baseline models were also implemented adapting their respective official codes \footnote{D-Persona official code: \url{https://github.com/ycwu1997/D-Persona}}\textsuperscript{,}\footnote{CIMD official code: \url{https://github.com/aimansnigdha/Ambiguous-Medical-Image-Segmentation-using-Diffusion-Models}}, and all default parameters were utilised. During inference with CIMD, the number of timesteps $T$ was reduced from 1000 to 100 to minimise the time required for inference, as it is significantly longer than that of other methods. All the probabilistic U-Nets (including the PULASki methods) required approximately 10 and 30 seconds per volume for MS and vessel segmentation, respectively. By contrast, the MH-VI U-Net (the second slowest baseline) required approximately 30 seconds and 6.5 minutes, respectively, for the same tasks. On the same hardware, CIMD, even with the reduced $T=100$, required over 1 hour for MS segmentation and more than 6 hours for vessel segmentation (with $T=1000$, it took more than 10 and 60 hours, respectively). Furthermore, as no stopping criterion for CIMD was specified in either the paper or the code, early stopping with a patience of 5 epochs for no change in the loss was implemented during model training. 

The models were trained by optimising the respective loss function for that particular model (as discussed earlier) using the Adam optimiser for 500 epochs. It is worth mentioning that all the models reached convergence before the pre-set number of epochs and the model state with the lowest validation loss was chosen for final evaluation. To reduce the training time per epoch, a limit of randomly chosen patch/slice was chosen with a uniform probability from all available samples, for each experiment -- 1500 and 4000, for 2D and 3D, respectively for the task of vessel segmentation, and 10000 for the task of MS segmentation. Increasing the limit will not have any impact on the results as all models converged before the 500th epoch -- rather, it will increase the time to finish 500 epochs, while converging after an earlier epoch. Trainings were performed using Nvidia 2080 TI, V100, A6000 GPUs -- chosen depending upon the computational demands of each model. Automatic mixed precision was employed for all the models, to reduce GPU memory requirements. The complete implementation of this project (PULASki method, as well as the baselines used here) is publicly available on GitHub\footnote{PULASki on GitHub: \url{https://github.com/soumickmj/PULASki}}, while the trained model weights are available on Hugging Face \footnote{PULASki weights on Hugging Face: \url{https://huggingface.co/collections/soumickmj/pulaski-66d9d35dfef91c84d140de8d}}.

It is worth mentioning that training 3D models is more challenging than their 2D counterparts, owing to the increase in computational complexity and overhead. Moreover, 3D experiments were not conducted for PULASki with FID loss, as it utilises a pretrained Inception Net V3, which is a 2D model. To employ FID loss in 3D, slices from the 3D patches must be considered separately, and the losses then need to be amalgamated, significantly increasing the computational requirements  and also rendering it not a purely 3D approach. Replacing the Inception Net V3 with an alternative 3D model could have been a possibility; however, this would no longer constitute the use of FID loss.



\subsection{Training and Evaluation}
\label{sec:traineval}

For training and evaluation of the network, the "StudyForrest" dataset was randomly divided into training, validation, and test sets in the ratio of 9:2:4. The MS dataset was divided accordingly and resulted in the ration 32:7:14 for training, validation and testing. Two sets of experiments were performed during this research: 2D and 3D. For the 2D experiments, 2D samples (commonly known as slices) were taken from the 3D volumes along the acquisition dimension (i.e. the slice-dimension that was used during the acquisition of the MRIs as they were all acquired in 2D and not 3D). On the other hand, 3D samples (commonly known as patches) of $64^3$ were taken from the volumes with dimensions, with strides of $32$, $32$, and $16$, across sagittal (width), coronal (height), and axial (depth), respectively. This was done to get overlapped samples and to allow the network to learn inter-patch continuity. During inference, the overlapped portions were averaged to obtain smoother output. It is to be noted that these sub-divisions of the available 3D volumes into 2D slices or 3D patches are essential to reduce the computational overheads, as well as to have more individual samples for the models during training. 



For evaluating the results, Generalised Energy Distance (GED) \cite{szekely2013energy,kohl2018} was employed, calculated using the following equation:
\begin{equation}
D_{\mathrm{GED}}^2\left(P_{\mathrm{Y}}, P_{\mathrm{S}}\right)=2 \mathbb{E}[d(s, y)]-\mathbb{E}\left[d\left(s, s^{\prime}\right)\right]-\mathbb{E}\left[d\left(y, y^{\prime}\right)\right],
\end{equation}
where $P_{\mathrm{Y}}$ is the ground-truth distribution, $P_{\mathrm{S}}$ is the distribution of the output of the model, $y$ and $y^{\prime}$ are sampled from $P_{\mathrm{Y}}$, $s$ and $s^{\prime}$ are sampled from $P_{\mathrm{S}}$, and finally $d$ is the distance measure 
, which in this research was the intersection over union (IoU) that can be calculated using:
\begin{equation}
    d = \frac{| \{i,j: s_{ij} = 1 \cap y_{ij} = 1 \} |}{| \{i,j: s_{ij} = 1 \cup y_{ij} = 1 \} |}
\end{equation}
where $y_{ij}$ and $s_{ij}$ denote the value at the pixel $(i,j)$, in ground-truth and prediction, respectively.  


Although GED serves as a metric to evaluate how close the distribution of the prediction is to the distribution of the annotations, it does not tell anything regarding whether or not the variability present in the annotation is similar to the variability in the predictions produced by the method. Krippendorff's alpha ($K\alpha$) \cite{krippendorff2018content} is a statistical measure that shows the agreement among the samples - commonly used to evaluate the inter-rater agreement. Hence, $K\alpha$ was used as an additional metric to evaluate how closely the methods reproduced the inter-rater disagreement, and is calculated as:

\begin{equation}
    K\alpha = 1 - \frac{D_o}{D_e}
\end{equation}
where $D_e$ is the expected disagreement assuming that raters are making their ratings randomly and $D_o$ represents the observed disagreement among raters calculated as:

\begin{equation}
    D_o = \frac{1}{N} \sum_{i=1}^{2} \sum_{j=1}^{2} o_{ij} \delta^2_{ij}
\end{equation}
where $N$ is the total number of pairwise comparisons between raters for each voxel, $o_{ij}$ is the observed frequency of voxels that are assigned to both category $i$ and $j$ in these pairwise comparisons, and $\delta^2_{ij}$ represents the difference function (0 for $i = j$ and 1 for $i \neq j$ for binary segmentation). For calculating $D_e$, $e_{ij}$ is used instead of $o_{ij}$, representing the expected frequency of voxels assigned to both category $i$ and $j$ if the raters were guessing without any systematic bias -- calculated by multiplying the probabilities (derived from marginal totals) of each category being chosen independently by two raters.  two distinct forms of $K\alpha$ values were computed: $K\alpha_{all}$ - $K\alpha$ calculations encompassing the entire volume and $K\alpha_{ROI}$ -  $K\alpha$ was computed solely based on the voxels identified by at least one rater. These two are collectively referred to as $K\alpha$ throughout this manuscript.  

Finally, the statistical significance of the variance noted across various methods and metrics was ascertained utilising the signed Wilcoxon Rank test \cite{wilcoxon1992individual}.

\section{Results}
\label{sec:res}



\subsection{Statistical measures of predicted segmentation variability} 
\label{sec:stat_seg_variab}
\begin{figure}[!h]
\centering
  \includegraphics[width=0.45\textwidth]{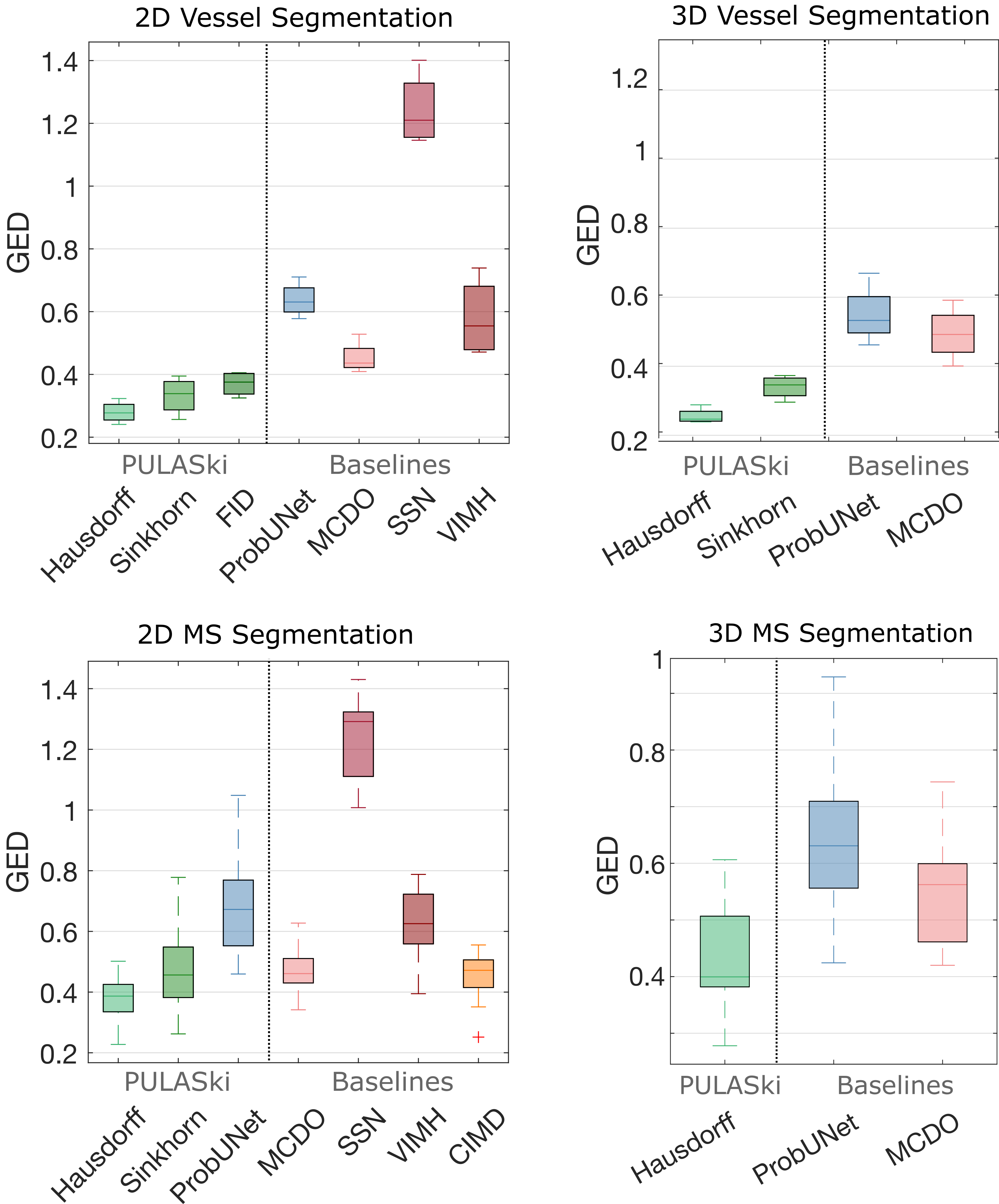}
  \caption{Quantitative assessment of distribution of generated segmentations per image compared to available data. Boxplots show variation in Generalised Energy Distance (GED) scores per image in the test set for all baselines and PULASKi with different statistical distances.  Results for the vessel and MS lesion segmentation task are shown in the top and bottom rows, respectively.}
  \label{Fig:GED_all}
\end{figure}

Evaluation metrics on predicted segmentations from the trained PULASKi show statistically significant improved representation of segmentation variability compared to all baselines.  These conclusions hold both when training on 2D slices and 3D patches and for both the vessel and MS lesion segmentation tasks. We evaluated Generalised Energy Distances (GEDs) and Krippendorff's alpha ($K\alpha$, see Section \ref{sec:traineval}) across 10 generated segmentations for a given image, over 4 images in vessel segmentation and 12 MS experiments respectively, as summarised in Figure \ref{Fig:GED_all} and Table \ref{tab:kalpha}.  The annotated data consists of 7 plausible annotations per image in MS and 10 in vessel. Boxplots in Figure \ref{Fig:GED_all} are are computed over the set of GED values calculated for each image.  Table \ref{tab:kalpha} shows $K\alpha$ values calculated over the entire volume ($\mathrm{K\alpha_{all}}$) and only on voxels classified as vessel or lesion in at least one annotation ($\mathrm{K\alpha_{ROI}}$). This second set of alpha values was chosen to minimise the influence of class imbalance, as only 0.13 \% and 2.56 \% of the overall volumes are lesions and vessels, respectively. It merits mention that the 2D PULASki FID and 3D PULASki Sinkhorn training failed to converge (i.e., got stuck in local minima), and hence were not included in the results. 


PULASKi with the Hausdorff divergence shows the best performance across both experiments, in terms of both GED and $K\alpha$.  It provides an improvement over the standard Probabilistic U-Net of 56 and 51 \% for 2D and 3D vessel segmentation and 46 and 33 \% for 2D and 3D MS lesion segmentation, in terms of the average GED.  The chosen statistical distances with PULASKi more often that not provide a statistically significant improvement in terms of both evaluation metrics over baselines at the 5\% significance level, particularly in the vessel segmentation task (see Table \ref{tab:p-vals}).  PULASKi (Hausdorff) was the only method to consistently outperform baselines at the 5\% significance level.  Amongst the baselines methods, MCDO with FTL loss allows for the best representation of inter-rater variability in terms of both GED (average of 0.45 for 2D and 0.49 for 3D in vessel segmentation and  0.47 for 2D and 0.55 for 3D in MS segmentation) and $K\alpha$, rather surprisingly.  VI-MH U-Net provides a slight improvement over the Probabilistic U-Net in the 2D experiments. The SSN was completely unable to represent the underlying variability in the training data, as evidenced by the large GED and Krippendorff alpha scores.  We posit that this is due to the cross-entropy term style loss function being bad for class imbalanced data and due to the lack of regularisation, which is present in PULASKi, Probabilistic U-Net and VI-MH U-Net. Both stages of D-Persona and CIMD with $T=100$ failed to produce consistent and reliable results for both MS and vessel segmentation tasks (quantitative results can be found in the supplementary materials, Fig. \ref{suppFig:GED_supp} and Tab. \ref{supptab:kalpha}). Hence, they were discarded from further analyses. Consequently, these methods were excluded from further analyses. For MS segmentation using CIMD, $T$ was increased to the original $T=1000$ (requiring more than 10 hours per volume), which yielded results comparable to those from MCDO in terms of GED (0.47). The low Krippendorff alpha scores indicate a high degree of heterogeneity in the generated plausible segmentations. The time required to process each volume (more than 60 hours) using CIMD with $T=1000$ for vessel segmentation rendered it impractical, and thus, this task was not undertaken. 

When evaluating difference in performance between 2D and 3D implementations, there is less agreement between GED trends and $K\alpha$ trends compared to overall performance between methods.  For instance, the PULASKi with Hausdorff shows improvements of about 20 \% in terms of $\mathrm{K\alpha_{ROI}}$ in the 2D implementation (23.63 and 31.28) compared to 3D (45.40 and 50.56) for both vessel and MS tasks, although the reverse is true with looking at median GED.  

Finally, we also assessed statistical significance of differences between the ground truth and predicted distributions from the PULASKi in terms of Krippendorff alpha scores.  Hausdorff 2D was the only method that did not have any statistically significant difference, as opposed to all other methods (including the baselines, p-values not shown in the table).  This suggests that while our PULASKi improves on existing methods, careful choice of the statistical distance is needed to ensure an accurate representation of inter-rater variability.

\begin{table}[!h]
\centering
\caption{Quantitative assessment of the distribution of generated segmentations using Krippendorff's alpha, calculated on the entire volume ($\mathrm{K\alpha_{all}}$) and only on voxels classified as vessel or lesion in at least one annotation ($\mathrm{K\alpha_{ROI}}$). 
 Values are indicated as mean over all images $\pm$ one standard deviation.  }
\label{tab:kalpha}
\begin{tabular}{@{}lcc@{}}
\toprule
\multicolumn{3}{c}{\textbf{Vessel Segmentation}}   \\ \midrule
&  \multicolumn{2}{c}{2D} \\  
& $\mathrm{K\alpha_{all}}$ & $\mathrm{K\alpha_{ROI}}$ \\ \midrule
Annotation & $40.24 \pm 8.62$  & $22.35 \pm 10.9$ \\ \midrule
PULASki HD & $50.34 \pm 2.31$ & $31.28 \pm 2.39$ \\
PULASki SH & $57.62 \pm 1.18$ & $34.96 \pm 1.3$ \\
PULASki FID & $50.21 \pm 0.39$ & $29.66 \pm 0.61$ \\
Prob UNet & $99.83 \pm 0.01$ & $63.08 \pm 0.59$ \\
MCDO UNet & $90.00 \pm 1.36$ & $58.71 \pm 1.07$ \\
SSN & $100.0 \pm 0$ & $100.0 \pm 0$ \\
VIMH & $88.78 \pm 1.79$ & $57.90 \pm 1.22$ \\ \midrule
&  \multicolumn{2}{c}{3D} \\ 
& $\mathrm{K\alpha_{all}}$ & $\mathrm{K\alpha_{ROI}}$  \\ \midrule
Annotation &  $40.24 \pm 8.62$ & $22.35 \pm 10.9$  \\ \midrule
PULASki HD  & $76.70 \pm 3.05$ & $50.56 \pm 3.18$ \\
PULASki SH & $82.85 \pm 2.27$ &  $53.30 \pm 2.36$ \\
Prob UNet  & $99.72 \pm 0.02$ & $62.74 \pm 0.17$ \\
MCDO UNet & $95.91 \pm 0.55$ &  $62.09  \pm 0.70$  \\
\toprule \bottomrule
\multicolumn{3}{c}{\textbf{MS Segmentation}}\\ \midrule
&  \multicolumn{2}{c}{2D} \\
& $\mathrm{K\alpha_{all}}$ & $\mathrm{K\alpha_{ROI}}$   \\ \midrule
Annotation & $60.40 \pm 14.06$ & $27.60 \pm 10.13$  \\ \midrule
PULASki HD & $62.06 \pm 11.88$  & $25.52 \pm 9.13$ \\
PULASki SH & $76.41 \pm 10.16$ & $40.75 \pm 8.85$  \\
Prob UNet & $99.93 \pm 0.03$ & $57.88 \pm 3.70$ \\
MCDO UNet & $82.93 \pm 8.19$  & $46.74 \pm 6.01$ \\
SSN & $100.0 \pm 0$ & $100.0 \pm 0$  \\
VIMH & $83.53 \pm 14.07$  & $45.78 \pm 10.13$  \\
CIMD & $36.54 \pm 18.48$  & $11.65 \pm 13.92$ \\ \midrule
&  \multicolumn{2}{c}{3D} \\ 
& $\mathrm{K\alpha_{all}}$ & $\mathrm{K\alpha_{ROI}}$  \\ \midrule
Annotation & $60.40 \pm 14.06$ & $27.60 \pm 10.13$  \\ \midrule
PULASki HD & $86.58 \pm 5.81$ & $49.00 \pm 4.31$  \\
Prob UNet & $99.92 \pm 0.03$ & $58.39 \pm 2.50$  \\
MCDO UNet  & $95.87 \pm 2.32$ &  $55.47 \pm 1.99$ \\
\bottomrule
\end{tabular}
\end{table}

\begin{table*}[]
\centering
\caption{$p$-values obtained using the Wilcoxon rank test while comparing the different PULASki methods against the baselines and annotations combining both data sets (*CIMD: only for MS Segmentation). Significance indicated at \( \alpha = 0.05 \) (\cmark: Significant, \xmark: Not Significant).} 
\label{tab:p-vals}
\resizebox{\textwidth}{!}{%
\begin{tabular}{llllllllllllllllll}
\cline{2-18}
& \multicolumn{3}{c}{Prob UNet} & \multicolumn{3}{c}{MCDO UNet} & \multicolumn{3}{c}{SSN} & \multicolumn{3}{c}{VIMH} & \multicolumn{3}{c}{CIMD*} & \multicolumn{2}{c}{Annotation} \\ \cline{2-18} 
\begin{tabular}[c]{@{}c@{}}PULASki \\ Method\end{tabular} & GED & \(\mathrm{K\alpha_{all}}\) & \(\mathrm{K\alpha_{ROI}}\) & GED & \(\mathrm{K\alpha_{all}}\) & \(\mathrm{K\alpha_{ROI}}\) & GED & \(\mathrm{K\alpha_{all}}\) & \(\mathrm{K\alpha_{ROI}}\) & GED & \(\mathrm{K\alpha_{all}}\) & \(\mathrm{K\alpha_{ROI}}\) & GED & \(\mathrm{K\alpha_{all}}\) & \(\mathrm{K\alpha_{ROI}}\)& \(\mathrm{K\alpha_{all}}\) & \(\mathrm{K\alpha_{ROI}}\) \\ \hline
Hausdorff 2D  & \cmark (\(<10^{-6}\)) & \cmark (\(<10^{-4}\)) & \cmark (\(<10^{-4}\)) & \cmark (\(<0.001\)) & \cmark (\(<10^{-4}\)) & \cmark (\(<10^{-4}\)) & \cmark (\(<10^{-7}\)) & \cmark (\(<10^{-4}\)) & \cmark (\(<10^{-4}\)) & \cmark (\(<10^{-6}\)) & \cmark (\(<10^{-4}\)) & \cmark (\(<10^{-4}\)) & \cmark (\(<0.05\)) & \cmark (\(<10^{-6}\)) & \cmark (\(<10^{-6}\)) & \xmark $(0.13)$ & \xmark $(1.0)$ \\
Sinkhorn 2D   & \cmark (\(<0.001\)) & \cmark (\(<10^{-4}\)) & \cmark (\(<10^{-4}\)) & \xmark $(1.0)$ & \cmark (\(<10^{-4}\)) & \cmark (\(<10^{-4}\)) & \cmark (\(<10^{-7}\)) & \cmark (\(<10^{-4}\)) & \cmark (\(<10^{-4}\)) & \cmark (\(<0.01\)) & \cmark (\(<10^{-4}\)) & \cmark (\(<0.001\)) & \xmark (\(0.8\)) & \cmark (\(<10^{-6}\)) & \cmark (\(<10^{-6}\)) & \cmark (\(<10^{-4}\)) & \cmark (\(<10^{-4}\)) \\
Hausdorff 3D  & \cmark (\(<10^{-5}\)) & \cmark (\(<10^{-4}\)) & \cmark (\(<10^{-4}\)) & \cmark (\(<0.01\)) & \cmark (\(<10^{-4}\)) & \cmark (\(<10^{-4}\)) & - & - & - & - & - & - & - & - & - & \cmark (\(<10^{-4}\)) & \cmark (\(<10^{-4}\)) \\ \hline
\end{tabular}%
}
\end{table*}

\subsection{Qualitative measures of predicted segmentation variability} 
\label{sec:qualmeas}
\begin{figure*}[htbp!]
\centering
\includegraphics[width=0.95\textwidth]{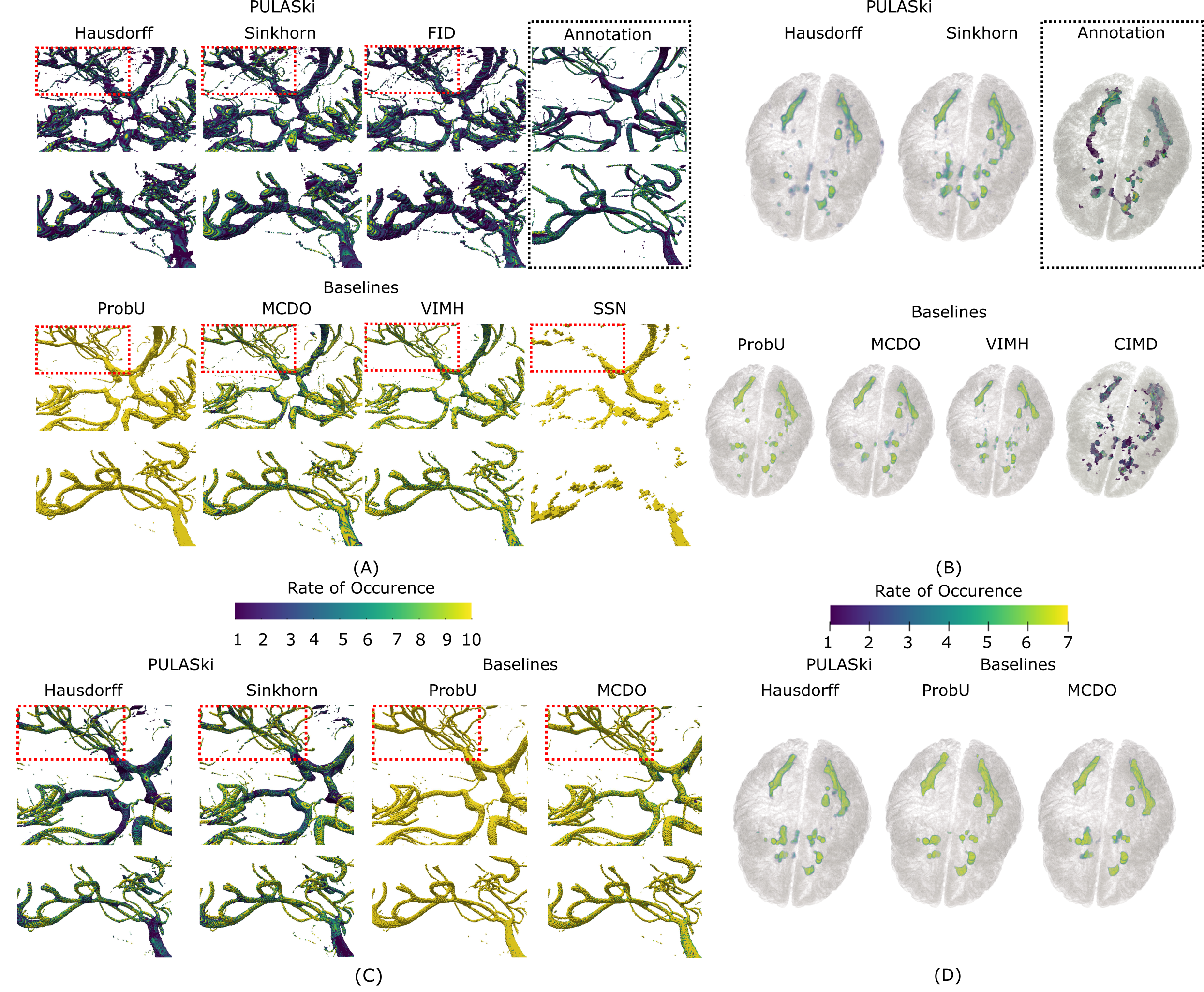}
  \caption{Rate of occurrence (RoO) of labels across generated segmentations for a given subject. The brighter (yellow) the voxel appears, the more often it
is labelled as vessel (A,C) or lesion (B,D).  The annotaded data is displayed in the dashed rectangle for vessel segmentation (A) and MS lesion (B). (A) RoO for the 2D vessel segmentation for PULASki method with different loss functions (FID, Sinkhorn and Hausdorff divergence) shown in the top row; baselines (Probabilistic U-Net, MCDO, VIMH, SSN) shown in bottom row. All methods were trained on 10 plausible labels per image. A detailed view of a specific region, indicated in the red rectangle, is provied below the larger volume.  (B) RoO  for the 2D Multiple sclerosis segmentation for PULASki method with different loss functions (Sinkhorn and Hausdorff) and selected baselines. All methods were trained on 7 plausible labels per image. (C) and (D) RoO for the 3D implementation in vessel segmentation and MS segmentation, respectively.}
    \label{Fig:RoO_all}
\end{figure*}

While evaluation metrics provide a concise quantitative measure of how closely the predicted distribution matches the annotations, qualitative results can be useful to determine which specific features are well represented by different methods. We plotted the frequency of labelling as vessel or lesion at a voxel level across 10 predicted segmentations from each method (Rate of Occurrence (RoO) see Figure \ref{Fig:RoO_all}. The brighter (yellow), the voxel appears, the more often it is labelled as a vessel or MS lesion throughout the output plausible segmentations. In line with the quantitative metrics, the PULASKi approach generates segmentations with structure and levels of agreement between annotations that are more consistent with the training data (see Figure \ref{Fig:RoO_OD}), for both vessel and lesion segmentation.  However, frequencies appear to be better represented in the vessel segmentation task than in the MS lesion task, where the extent of lesions is smaller and level of agreement across predicted segmentations higher than in the training data (see Figure \ref{Fig:RoO_OD}(B)).  This is true not just for PULASKi, but across all methods and could be attributed to the lower inter-rater agreement in the annotated MS data.  Finally, vessel thickness and lesion extents are slightly over-estimated in the PULASKi method in 2D slice based training compared to 3D volumes. CIMD, as a diffusion-based model, produces segmentations that exhibit the highest heterogeneity compared to all other models. However, in comparison to the annotations, it identifies regions as lesions that were not selected by the annotators, which can possibly be attributed to the issue of "hallucination of lesions".

The baseline probabilistic U-Net provides reasonable segmentations in terms of vessel structure and MS lesion shape, however it is clear from Figure \ref{Fig:RoO_all} that the variability between segmentations is very poorly represented.  Both MCDO and VI-MH U-Net demonstrate superior performance to the probabilistic U-Net; the improvement of the VI-MH over the probabilistic U-Net is much clearer in Figure \ref{Fig:RoO_all} compared to the evaluation metrics where there was not a statistically significant difference. Consistent with the evaluation metrics, the SSN provides unrealistic, disjointed vessel segmentations and poor representation of variability. Smaller vessels are particularly poorly segmented in the SSN, suggesting 
that the DeepMedic architecture might not be suitable for small vessel segmentation. But the multi-scale nature of DeepMedic should have been better for large, as well as small vessels. But maybe it simply ignored the small vessels and focused only on the large vessels. The dual-pathway idea of DeepMedic might have been acting against vessels that appeared only on one pathway (higher scale).  In fact, methods that are more capable of producing segmentations with less agreement between annotations are better able to segment small vessels (see red rectangle in Figure \ref{Fig:RoO_all}), although many of these vessels are unrealistically disjointed. This indicates that further refinements are needed to simultaneously segment both large and small vessels, whilst representing underlying variability in annotations.

\subsection{Pixel-wise Probability Distribution} 
\begin{figure}[!h]
\centering
\includegraphics[width=0.46\textwidth]{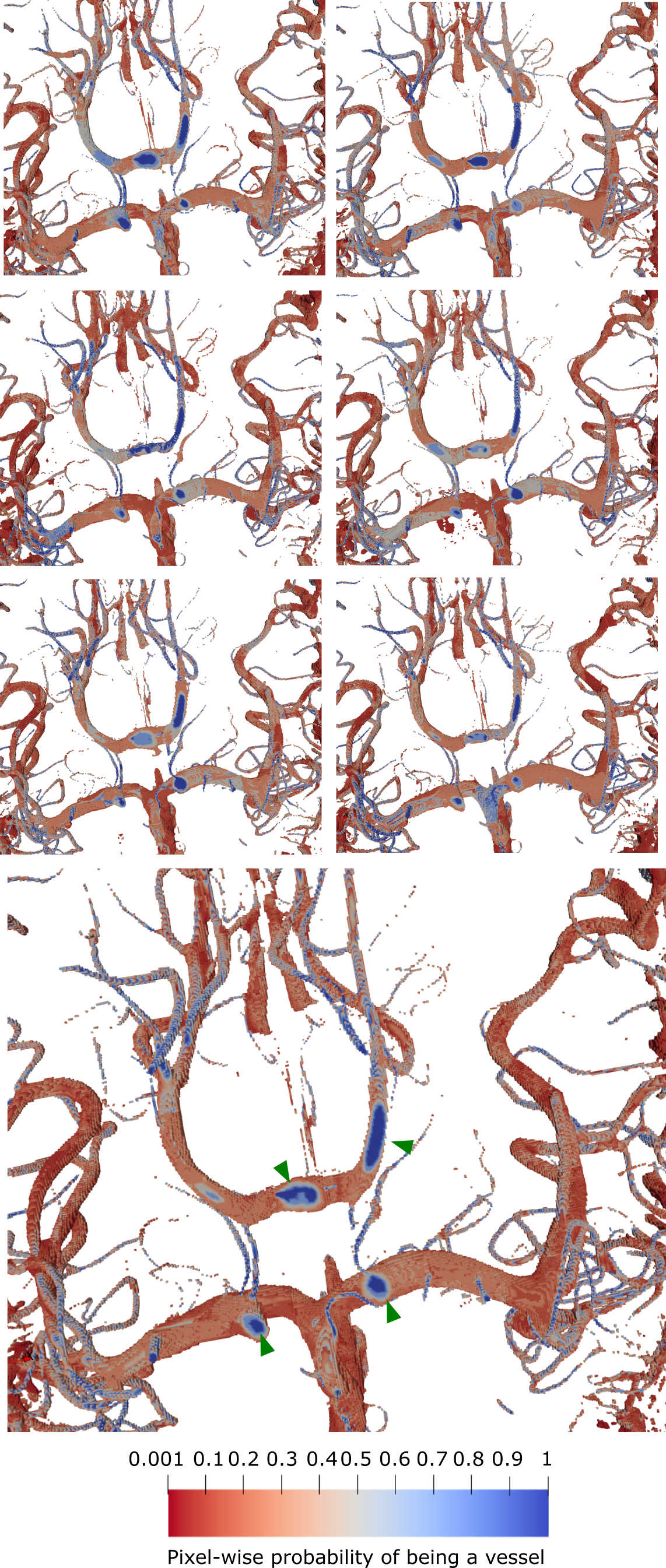}
  \caption{3D volume of pixel-wise probability of being a vessel for 6 (exemplary chosen) of the 10 generated plausible segmentations in the 3D Hausdorff implementation. The median over the 10 output segmentations is displayed at the bottom.  Higher probability of the voxel being a vessel is shown in blue, while red indicates low probability. The green arrows indicate internal vessel parts that become visible because the measurement field of view ends at that specific location.}
    \label{Fig:3D_RawProbPred}
\end{figure}
Pixel-wise class probabilities, or marginal probabilities, are useful for a variety of downstream tasks although they only partially capture the dependence structure between pixels.  Such tasks include setting appropriate thresholds to estimate vessel diameter in surface mesh generation, or to estimated uncertainty of hemodynamic simulations in underlying surface meshes. Here we consider pixel-wise probabilities to be the output of the soft-max operator (i.e. the logits), noting that the probability itself is random in all methods considered. Different to the RoO discussed in Section \ref{sec:qualmeas}, variations in the pixel-wise probabilities can be used as a measure of model uncertainty.   As evident in Figure \ref{Fig:3D_RawProbPred}, there can often be significant variability in pixel-wise class probabilities at vessel junctions and boundaries.  This holds for the PULASKi with Hausdorff divergence; similar levels of variability were not observed for the baselines method consistent with the under-estimation of uncertainty in terms of other metrics.  Conversely, pixel-wise probabilities are much less variable along the vessel centreline and are closer to probability 1 than at other locations as expected.

\begin{figure*}[htbp!]
    \centering
  \includegraphics[width=15.7cm,height=8.5cm]{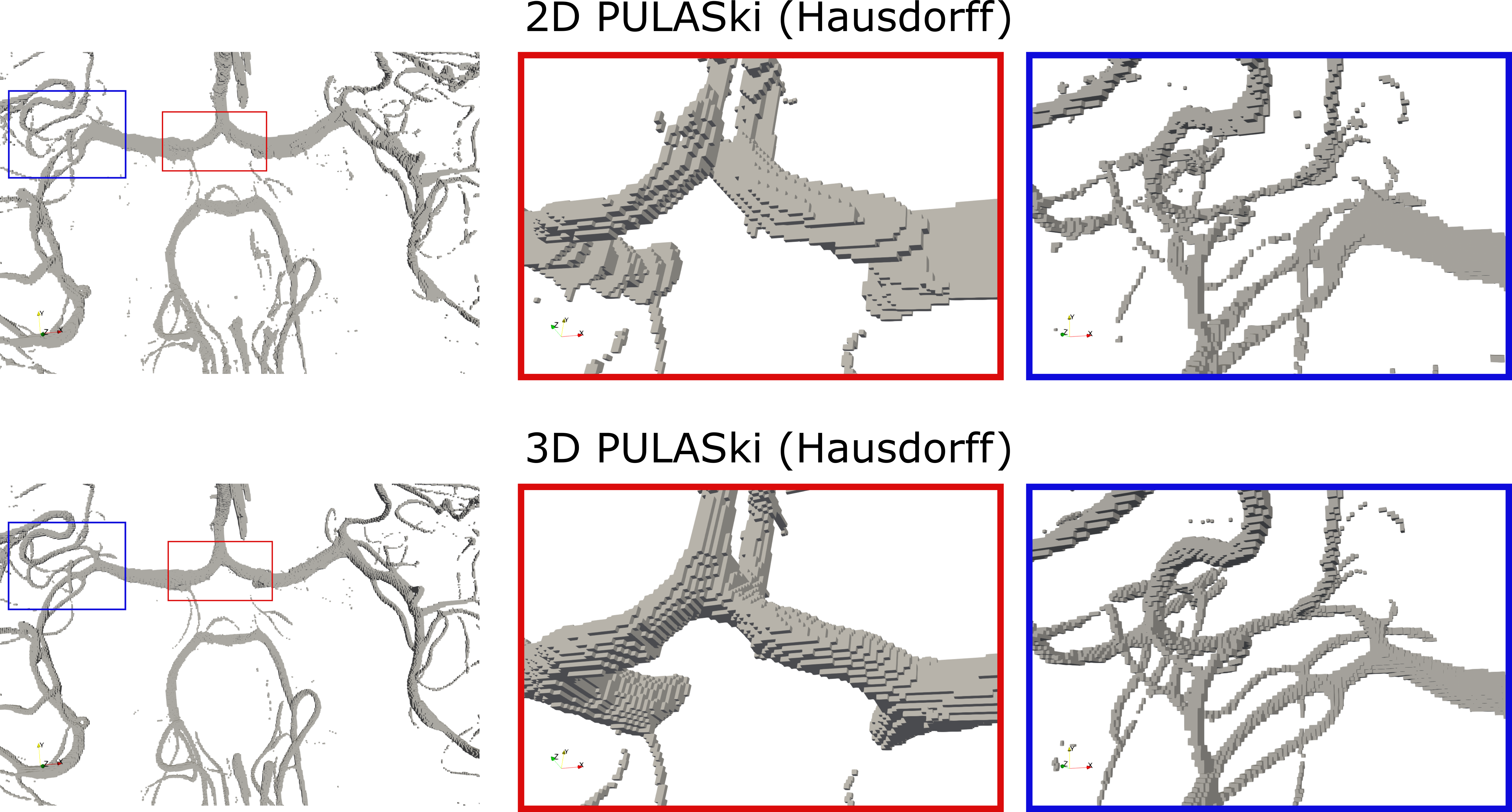}
  \caption{Comparison of 2D and 3D vessel segmentation for the PULASki method with Hausdorff divergence, exemplary shown for one plausible generated segmentation. The full volume is shown in the left column. Detailed views are displayed showing larger vessels (red square, middle column) and smaller vessels (blue square, right column), respectively.}
    \label{Fig:2D_3D_comp_v2}
\end{figure*}

\subsection{2D vs 3D models} 
\label{subsec:2D_3D}
\begin{figure}[htbp!]
\centering
\includegraphics[width=0.45\textwidth]{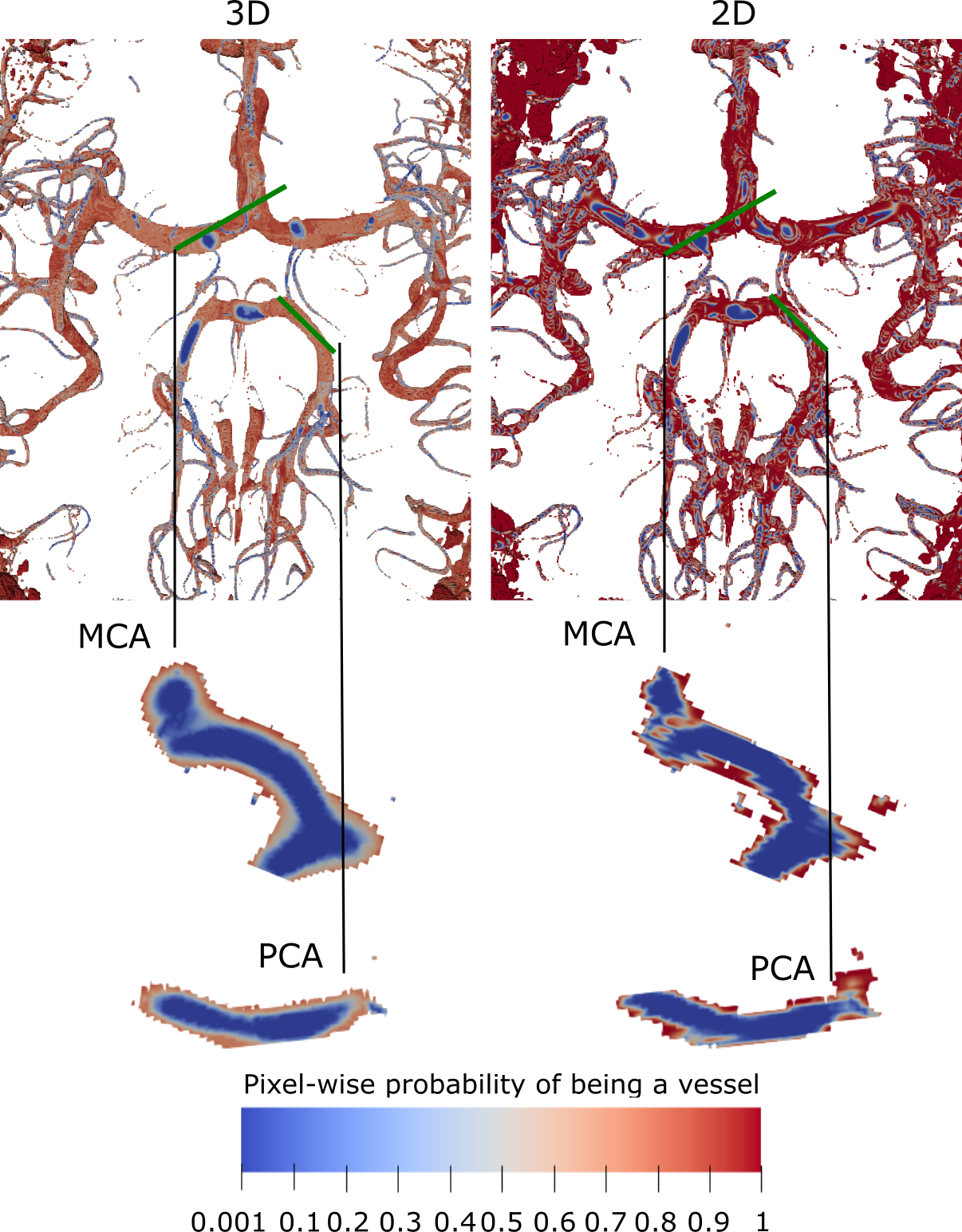}
  \caption{Comparison of pixel wise probability maps (median over generated segmentations) for a given subject in 2D vs 3D Hausdorff implementation for the vessel segmentation task. The median was calculated over probability maps corresponding to 10 generated segmentations from the trained PULASKi model.  The whole volume is displayed in the top row and cross sections of the middle and posterior cerebral arteries (MCA and PCA) are presented in the bottom row, with corresponding locations indicated by green lines.}
    \label{Fig:3D2D_RawProbPred}
\end{figure}
\begin{figure}[htbp!]
\centering
\includegraphics[width=0.45\textwidth]{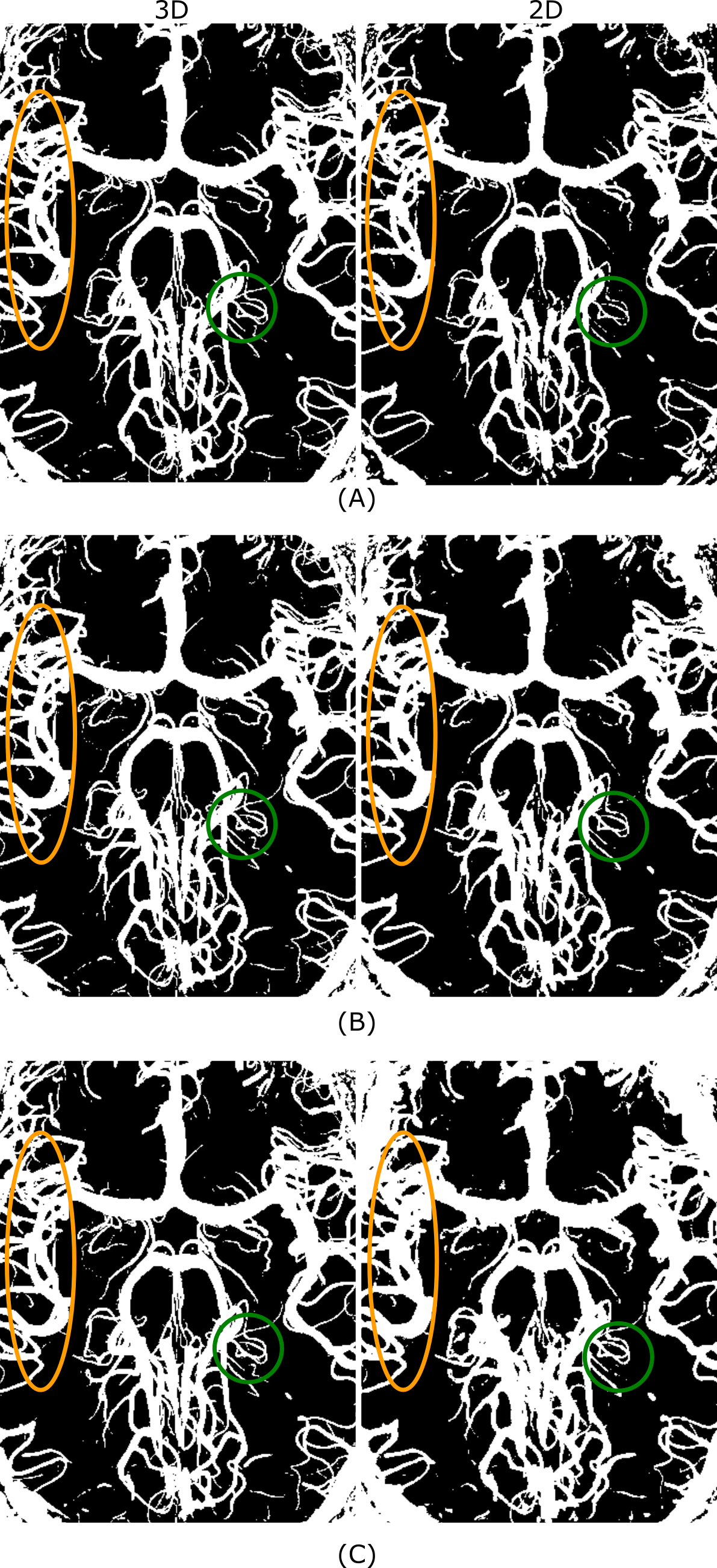}
  \caption{Comparison Maximum Intensity Projection (MIP) generated from the median pixel-wise probability maps for a given subject in 2D vs 3D Hausdorff implementation for the vessel segmentation task. Results are shown for threshold of 0.1 in (A), 0.01 in (B) and 0.001 in (C). The green and orange circle encompasses a region that includes small vessels, and its sensitivity to selected thresholds is particularly pronounced in the 2D implementation.}   
    \label{Fig:3D2D_RawProbPred2}
\end{figure}

Training models to segment inherently 3D volumes using training data consisting of 2D slices, whilst computationally cheaper, generally produces poorer quality segmentations. As described in Section \ref{sec:stat_seg_variab}, GED scores were comparable between 2D and 3D implementations of the best performing method (PULASki with Hausforff distance), while the $K\alpha$ value was slightly lower for 3D compared to 2D.  However, 3D mesh visualisation of the segmented vessels  (Figure \ref{Fig:2D_3D_comp_v2}) reveals unrealistic ``stacked'' or ``stapled'' regions more prevalent in the 2D implementation than 3D. 
Moreover, it can be observed in Figure \ref{Fig:3D2D_RawProbPred} that the pixel-wise probability of a voxel being a vessel exhibits a nearly discrete distribution of probability in 2D, whereas in 3D it gradually decreases from the centre to the edges.
This can be attributed the lack of continuity between 2D slices, which is much better preserved when training on 3D patches. In the 3D implementation, predictions during inference are generated in overlapping patches. The final prediction is obtained by averaging over the overlapping region, thereby providing a better transition across patches. This makes the 3D results better suited for downstream tasks, e.g., 3D surface generation for CFD or DA, where smooth transitions are important. This smoothness can further be observed in figure \ref{Fig:3D_RawProbPred}, where pixel-wise probabilities have a much more spatially consistent structure in 3D than 2D. However, the averaging operation that allows smoother transitions in the 3D implementation has the potential to reduce variability in the generated segmentations, as indicated in Figure \ref{Fig:RoO_all} and the higher $K\alpha$ values (Table \ref{tab:kalpha}). 
Moreover, the 3D implementation yields more certain pixel-wise probabilities. The Maximum Intensity Projection (MIP) of the pixel-wise probabilities for all values that exceed a probability of 0.1 (A), 0.01 (B) and 0.001 (C) confirm greater certainty and stability in the 3D implementation (Figure \ref{Fig:3D2D_RawProbPred2}). Specifically, for small vessels (green and orange circle), the 2D projection represents the area only for the lower threshold as a connected vessel segment, whereas the 3D implementation illustrates analogous structures for both thresholds. Although smaller vessels appear adequately represented in the lower threshold of the 2D segmentation, larger vessels appear to be overestimated. This is particularly evident when considering the large vessels in the 0.001 threshold. They appear unrealistically large and overestimated while obscured by noise.

\subsection{Anatomical plausibility of generated segmentations} 

\begin{figure*}[htbp!]
\centering
\includegraphics[width=0.95\textwidth]{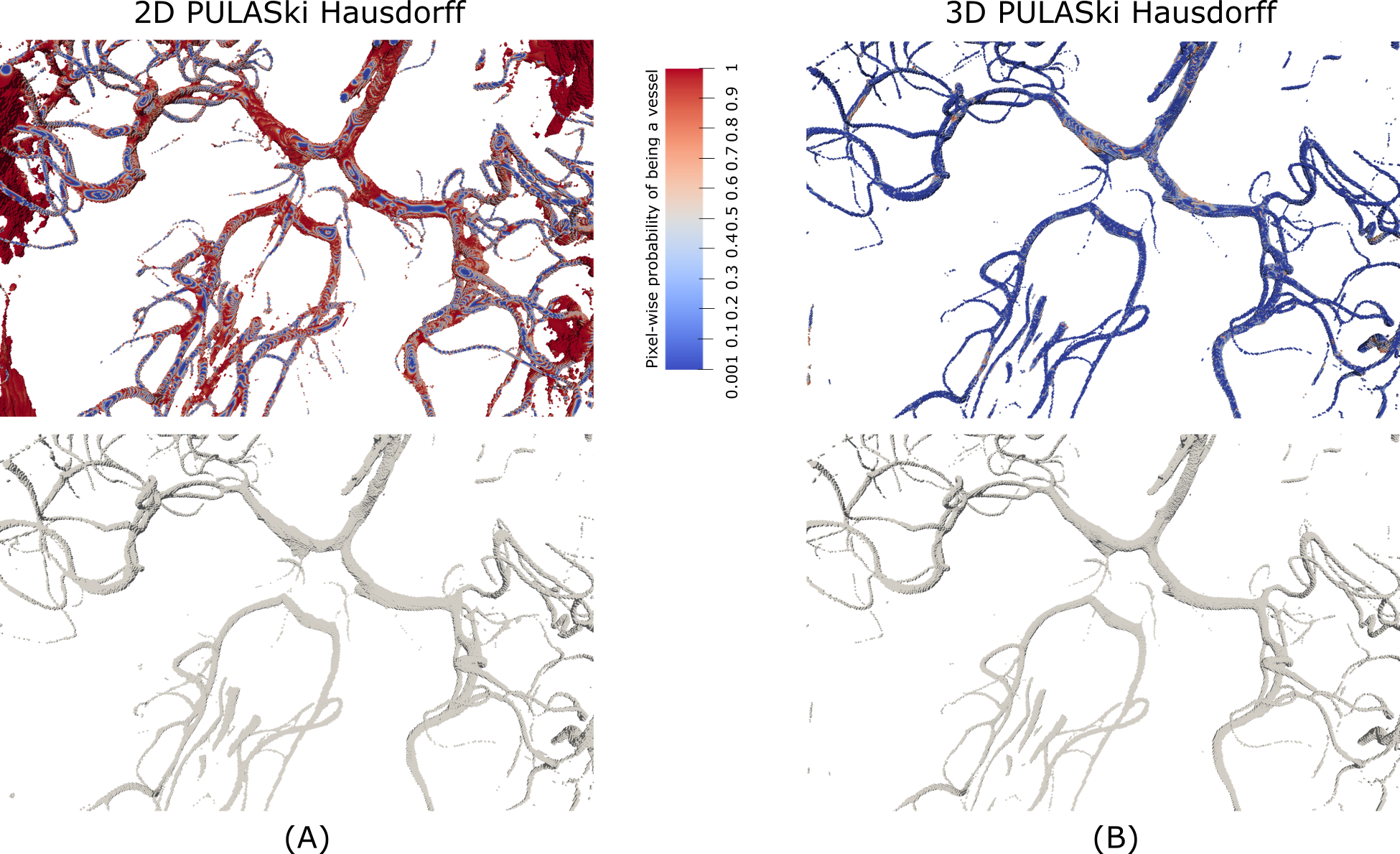}
  \caption{Approximation of the most probable (maximum likelihood) segmentation from the PULASKi method in 2D (A) and 3D (B) implementation.  This was calculated by adopting the prior mean for the latent vector $z$.  The top row shows the corresponding pixel wise probabilities of being a vessel and the bottom row displays the final segmentation, obtained by applying Otsu's method to threshold the probability map.} 
    \label{Fig:AnaPlaus}
\end{figure*}

\begin{figure*}[htbp!]
\centering
\includegraphics[width=0.95\textwidth]{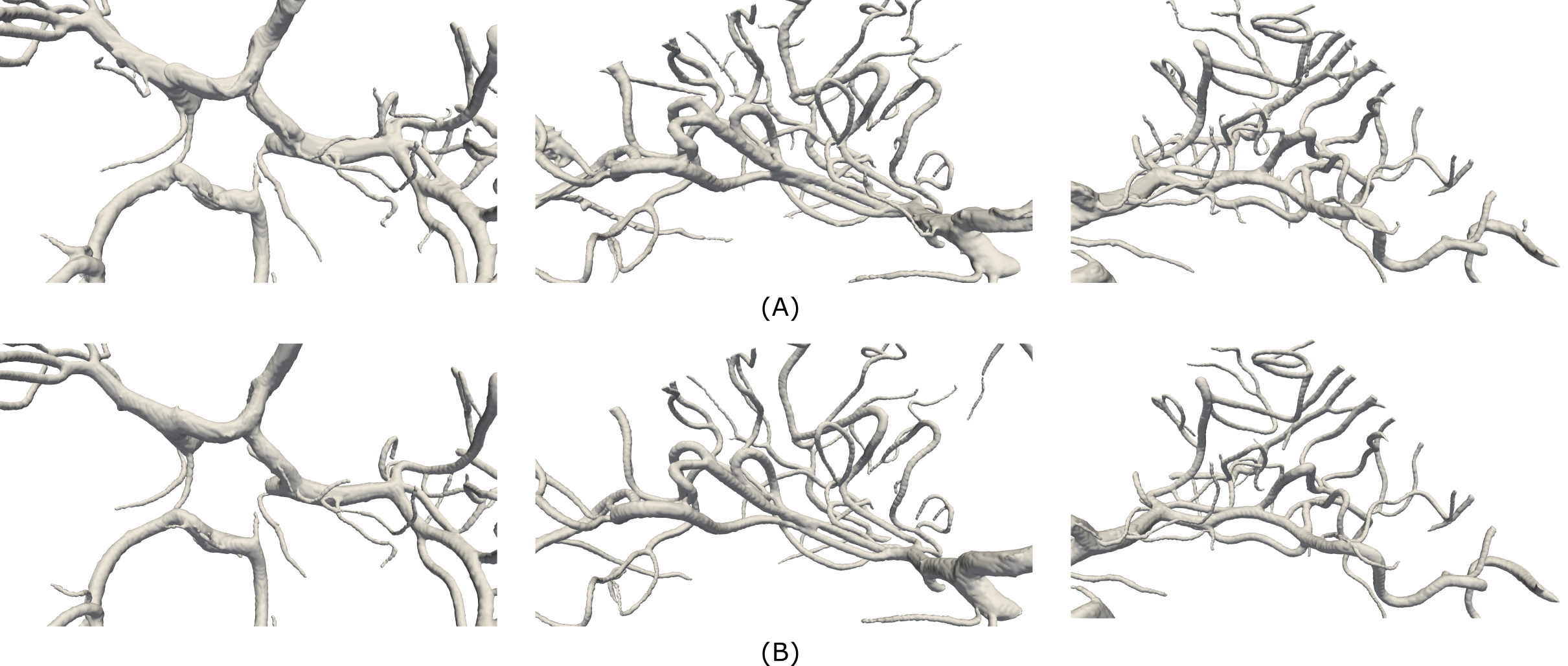}
  \caption{Surface mesh created using the most probable segmentation from the 2D (A) and 3D (B) Hausdorff PULASki implementation. The CoW region is depicted in the left column showcasing larger vessels, while the middle and right columns display zoomed views of smaller vessel structures close to the right and left middle cerebral arteries.} 
    \label{Fig:AnaPlaus_2}
\end{figure*}

Several downstream tasks necessitate not only an estimation of the most likely segmentation, but also the generation of smooth surface meshes for the segmented volume. For instance, conducting a CFD simulation within a segmented vessel demands seamlessly connected and smooth surface structures. Figure \ref{Fig:AnaPlaus} shows pixel-wise probabilities (logits) generated from the PULASKi (Hausdorff) with the prior mean of the latent variable $z$.   
This was then used together with Otsu's method to generate an approximation of the most probable segmentation (Figure \ref{Fig:AnaPlaus_2}).  Both 2D and 3D implementations yield segmentations that are in large part anatomically credible, in particular the Circle of Willis region is showing connected anterior, posterior and middle cerebral arteries. The left posterior communicating artery (Pcom) is reconstructed with the 3D implementaion, while the right Pcom is disconnected. However, this could also be attributed to the individual anatomy of the CoW in individual patients, which can be a physiological normal variant 
The most probable segmentation also appears to be significantly more anatomically credible (in terms of less stacked/stapled segmentations, disjointed vessels and smoother tube like vessel structures) than the randomly generated segmentations (see Figure \ref{Fig:2D_3D_comp_v2}). 
Most notably, Figure \ref{Fig:AnaPlaus} demonstrates increased pixel-wise uncertainty in the 2D implementation relative to its 3D counterpart. This could be attributed to the fact that in 2D, the model learns and infers in a disjoint fashion (i.e., each slice is learnt and predicted independently), and also the models do not learn the relation in the $Z$ direction. While in 3D, the model learns from overlapping 3D patches, and during inference, overlapped parts of the patches are averaged -- parts of the images with less certainty are averaged out, resulting in more confident segmentations. 
This discrepancy in 2D has implications for the generation of surface meshes necessary for CFD and other numerical simulations, as illustrated in Figure \ref{Fig:AnaPlaus_2}. As anticipated, the surface structure of the meshes is smoother in the 3D implementation (especially for the larger CoW region).  Additionally, the 2D implementation has a tendency to produce thicker vessel segmentations (Figure \ref{Fig:AnaPlaus_2}) than in 3D which is in line with the higher pixel-wise uncertainty and likely due to inefficiencies of the 2D implementation rather than reflecting actual vessel thickness. However, the 2D meshes are satisfactory and could be utilised for downstream tasks with additional processing steps, such as smoothing. This may be desirable in cases where the computational overhead of a 3D implementation for training is prohibitive.


\subsection{Methodological insights} 
\begin{table}
\centering
\caption{Comparison of the network architectures in terms of the number of trainable parameters}
\label{tab:train_params}
\begin{tabular}{@{}lcc@{}}
\toprule
                                                                                  & 2D        & 3D         \\ \midrule
\begin{tabular}[c]{@{}l@{}}Probabilistic UNets\\ (including PULASki)\end{tabular} & 7,509,580 & 22,512,940 \\
MCDO UNet                                                                         & 7,701,825 & 22,399,425 \\
SSN                                                                               & 425,287   & 1,042,507  \\
VIMH                                                                              & 1,899,968 & 5,519,360  \\ \bottomrule
\end{tabular}
\end{table}

\begin{table}[]
\centering
\caption{Comparison of the computational complexity of the models by the means of MACs}
\label{tab:model_macs}
\begin{tabular}{@{}lccc@{}}
\toprule
                                                                                  & \multicolumn{3}{c}{MACs ($\times10^9$)}                                                                                                                                  \\ \midrule
                                                                                  & \begin{tabular}[c]{@{}c@{}}2D\\ VSeg\end{tabular} & \begin{tabular}[c]{@{}c@{}}2D\\ MSSeg\end{tabular} & \begin{tabular}[c]{@{}c@{}}3D\\ VSeg/MSSeg\end{tabular} \\ \midrule
\begin{tabular}[c]{@{}l@{}}Probabilistic UNets\\ (including PULASki)\end{tabular} & 54.75                                             & 11.51                                              & 58.61                                                   \\
MCDO UNet                                                                         & 92.70                                             & 19.48                                              & 97.63                                                   \\
SSN                                                                               & 34.93                                             & 7.34                                               & 113.70                                                  \\
VIMH                                                                              & 26.61                                             & 4.32                                               & 45.79                                                   \\ \bottomrule
\end{tabular}
\end{table}


\begin{table}[htbp!]
\centering
\caption{Comparison of model efficiency, in terms of MACs (of the network architectures) per trainable parameter, presented in logarithmic (base 10) scale}
\label{tab:macs_per_param}
\begin{tabular}{@{}lccc@{}}
\toprule
& \multicolumn{3}{c}{MACs/Parameter (log)} \\ 
\midrule & \begin{tabular}[c]{@{}c@{}}2D\\ VSeg\end{tabular} & \begin{tabular}[c]{@{}c@{}}2D\\ MSSeg\end{tabular} & \begin{tabular}[c]{@{}c@{}}3D\\ VSeg/MSSeg\end{tabular} \\ \midrule
\begin{tabular}[c]{@{}l@{}}Probabilistic UNets\\ (including PULASki)\end{tabular} & 3.86 & 3.19 & 3.42 \\
MCDO UNet & 4.08 & 3.40 & 3.64 \\
SSN & 4.91 & 4.24 & 5.04 \\
VIMH & 4.15 & 3.36 & 3.92 \\ \bottomrule
\end{tabular}
\end{table}

The superior performance of PULASKi with an appropriate statistical distance compared to baselines can be primarily attributed to two factors; 1) effective learning of a low dimensional representation on which uncertainty quantification can be performed efficiently and 2) the chosen distribution based reconstruction losses that better capture inter grader variability (that is, to better learn $p(y|z,x)$).  A significant challenge in the case studies considered here is class imbalance; for the MS training data only 0.13\% of the full training data set corresponds to lesions and 2.56\% corresponds to vessel in the intracranial vessel segmentation data on average. 
PULASKi is more robust to class imbalance issues than the adopted baselines, as no specific measures were adopted to deal with it while the original formulation of the Probabilistic U-Net failed to even converge during training until the cross-entropy was replaced with the Focal Tversky Loss (FTL) \cite{abraham2019novel}. 

The chosen statistical distances in PULASKi are also numerically better suited to comparing the distribution of segmentations \textit{for a given image} with the model output distribution.  Empirical cross-entropy style losses used in the baselines poorly capture the variability in the conditional distribution $p(y|x)$, as the log probabilities for all segmentations, regardless of the corresponding image, are lumped together through a sum (see e.g., \ref{eq:probunetloss}).  Poor performance of the SSN can be attributed to similar issues, given the similarity of the loss function to that of the Probabilistic U-net, although there are likely additional shortcomings of the Deepmedic architecture for the tasks at hand. The lower resolution input might ignore smaller structures - a very frequent occurrence in both tasks performed here. Thed  We attribute improved performance of the PULASKi using the Hausdorff divergence, over the theoretically more accurate de-biased Sinkhorn divergence to potential instabilities introduced when solving for the more complicated optimal transport loss in (\ref{eq:sinkhorn}).  The use of multiple ``heads'' in the MH VI-Unet appears to not be flexible enough to represent inter-rater variability in the examined case studies, although this could potentially be improved by increasing the number of heads.  This would also then increase the computational expense. The latest baseline employed in this work, CIMD (with $T=1000$), yielded a comparable GED to the MCDO U-Net, whilst exhibiting heterogeneity exceeding that of the ground truth for the 2D MS segmentation task. However, the inference time of this model, being significantly greater than that of all other methods, coupled with the issue of "hallucination of lesions" identified through closer qualitative analysis, rendered the model neither particularly useful nor reliable in comparison to the others. Owing to the exceedingly high inference time, CIMD (with $T=1000$) was not utilised for 2D vessel segmentation or either of the 3D segmentation tasks. 

Efficiency of the network architecture as well as accuracy of automatic segmentation methods are important, particularly when used with systems with limited hardware capabilities as often found in clinical settings. 
Efficiency with respect to the number of trainable parameters and computational complexity can be assessed by the Multiply-Accumulate Operations (MACs) per trainable parameter. To determine this we first calculated the MACs for a forward pass of each architecture and can be found in Table \ref{tab:model_macs}. Since MACs depend on input data size, there are differences between the vessel and lesion segmentation task in the 2D implementation as the input slices differ in size. However, the 3D implementation relies on patches of the same size in both tasks (see Table \ref{tab:model_macs}).  The total number of trainable parameters was also computed for each architecture (Table \ref{tab:train_params}), which also helps provide some insights as to the differing accuracy of the methods. The significantly lower number of trainable parameters in the SSN (with DeepMedic) could indicate that the architecture is not sufficiently flexible for the tasks at hand, to represent complex segmentations. 
The better performing methods (MCDO, PULASKi) have a similar number of trainable parameters, significantly more than in the SSN.  In terms of efficiency, we see that the Probabilistic U-Net based methods (including PULASKi) are most efficient (Table \ref{tab:macs_per_param}), with SSN the least efficient despite the very low number of trainable parameters.  However, it is worth mentioning that during training, there are additional overheads (e.g., loss calculation and backward pass), and the best performing distances (Hausdorff and Sinkhorn) are computationally more expensive than evaluating the cross-entropy.  

\section{Discussion} 
\label{sec:discussion}


Segmenting complex vessel geometries and tissue lesions from noisy MRI remains a challenging task, despite improvements in imaging technology.  Often times there is significant variation in labels from human annotators for the same image. Supervised and semi-supervised deep learning based segmentation is rapidly developing to provide more accurate results in a fraction of the time of traditional methods. Ensuring these methods faithfully capture the underlying uncertainty in labelled data is paramount, so as to prevent over-confident diagnoses of diseases and model predictions.  Our method provides a significant improvement on many existing methods for capturing inter-rater variability in deep learning based medical image segmentation, for the same amount of training data.  It is easily implementable and can be used to provide practitioners with pixel-wise uncertainties and generate several whole volume plausible segmentations for Monte Carlo-based modelling purposes involving computational fluid dynamics models and data assimilation \cite{gaidzik2021hemodynamic}.

Experimental results on intracranial image segmentation tasks show that the distribution based loss functions utilised in our proposed method, PULASKi, drastically improve the ability to learn diverse plausible segmentations for a given image.  The loss functions are also far less sensitive to class imbalance in the training data set, based on experiments in both vessel and multiple sclerosis segmentation tasks in 2D and 3D cases.  Furthermore, there are minimal extra hyper-parameters to be specified compared to existing methods (e.g., probabilistic U-net), whilst simultaneously providing dramatic improvement in performance.  Our experimental results also demonstrate a significant improvement in both the anatomical plausibility of generated segmentations and representation of uncertainty when training on 3D patches rather than 2D slices.  In particular, vessel thickness seems to be overestimated easily in the 2D implementation with an uncertain threshold and vessel boundaries are not as smooth as in 3D. In terms of practical applications, presumably 3D implementations are generally more beneficial, particularly when smooth, continuous surfaces are required for advanced downstream tasks, such as in CFD. While 2D approaches can still be useful for gaining an initial understanding of the distribution of possible segmentations, they are not ideal for tasks that demand high accuracy and reliability. In the medical field, 2D segmentation may be sufficient where the focus is on identifying potential regions - providing a general idea - where lesions (e.g. MS or tumour) could occur, rather than needing to delineate smooth, continuous boundaries. 

It is to be noted that training in 3D significantly increases the computational requirements. Inference using 3D patches (like done here) might result in some reduction in variability due to the averaging operation performed for the overlaps (as discussed earlier). This can possibly be avoided by performing the inference on the whole 3D volume. Given the computational limitations, we could not perform it during this research.


The representation of small vessels is drastically improved using 3D training data, a task that has traditionally been incredibly challenging. Manual and semi-automated segmentation, despite being time-consuming, is typically deemed reliable primarily for medium to large vessels, as indicated in previous studies \cite{jerman2015, Chatterjee2020}. The proposed work is one of the first studies to demonstrate reliable segmentation of small vessels and supports the results of \cite{Chatterjee2020} by producing a set of plausible segmentations. Moreover, PULASki was the only method showing variability in the subset of possible segmentations within the area of smaller vessels, as illustrated in the RoO results (Figure \ref{Fig:RoO_all}) This also aligns more closely with the annotated data, as depicted in \ref{Fig:RoO_OD}. This provides significant opportunity to improve the modelling of cerebral small-vessel which supports the research in diseases that are difficult to predict and manage, e.g., CSVD. In the MS segmentation task, PULASKi also provides an improved representation of inter-rater variability over the baselines, although ``outlier'' or ``extreme'' annotations are still not represented well. This suggests potential for further improvements to quantify distributional tails more accurately.  

Although the focus of our work is in providing multiple plausible segmentations, there can be situations where obtaining one single segmentation per input is required (e.g. at clinics). In such scenarios, majority voting from the predicted plausible segmentations can be taken, or even the segmentation from the prior mean can be utilised.

It should be noted that the distribution-based loss functions in PULASKi are more computationally expensive to evaluate than the standard cross-entropy or FTL used in the benchmark methods.  However, we anticipate that this is generally not prohibitive, as training is typically undertaken offline and there is no further computational burden for inference tasks, compared to other variational autoencoder based methods. It is also to be noted that this research did not use FID loss for the 3D experiments as the loss model was 2D and this can be extended in the future by using a 3D loss calculation model trained on a 3D dataset.

Further challenges with PULASki include its ability to adequately capture the full range of distribution for both vessel and multiple sclerosis segmentation. Whilst the 3D segmentation is more robust, it nevertheless fails to encompass the entire distribution range and may omit certain outliers. However, it demonstrates closer alignment with the annotated range compared to the baseline models we evaluated.

Moreover, the current implementation can erroneously identify regions devoid of vessels as containing vessels. To mitigate this issue, we propose incorporating a consistency criterion to evaluate physical plausibility in future work. This approach would be contingent upon the specific task for which PULASki is employed; for instance, in vessel detection, the code could assess connectivity to ensure accuracy. These observations underscore areas where our implementation could be refined to improve its efficacy in segmentation tasks.

\section*{Acknowledgements}
This work was in part conducted within the context of the
International Graduate School MEMoRIAL at Otto von Guericke
University (OvGU) Magdeburg, Germany, kindly supported by the
European Structural and Investment Funds (ESF) under the program
“Sachsen-Anhalt WISSENSCHAFT Internationalisierung” (project
no. ZS/2016/08/80646). This work was further supported by the Ministry
of Economics, Science and Digitization of Saxony-Anhalt in
Germany within the Forschungscampus STIMULATE (grant number
I 117). HM was supported in part by the German Research Foundation (DFG) project number MA 9235/1-1 (446268581) and MA 9235/3-1 (501214112) as well as by the Deutsche Alzheimer Gesellschaft (DAG) e.V. (MD-DARS project).  This research has been partially funded by Deutsche Forschungsgemeinschaft (DFG)- SFB1294/1 - 318763901. 

\bibliography{mybibfile}

 \section*{Supplementary}
 \begin{figure}[!h]
\centering
  \includegraphics[width=0.45\textwidth]{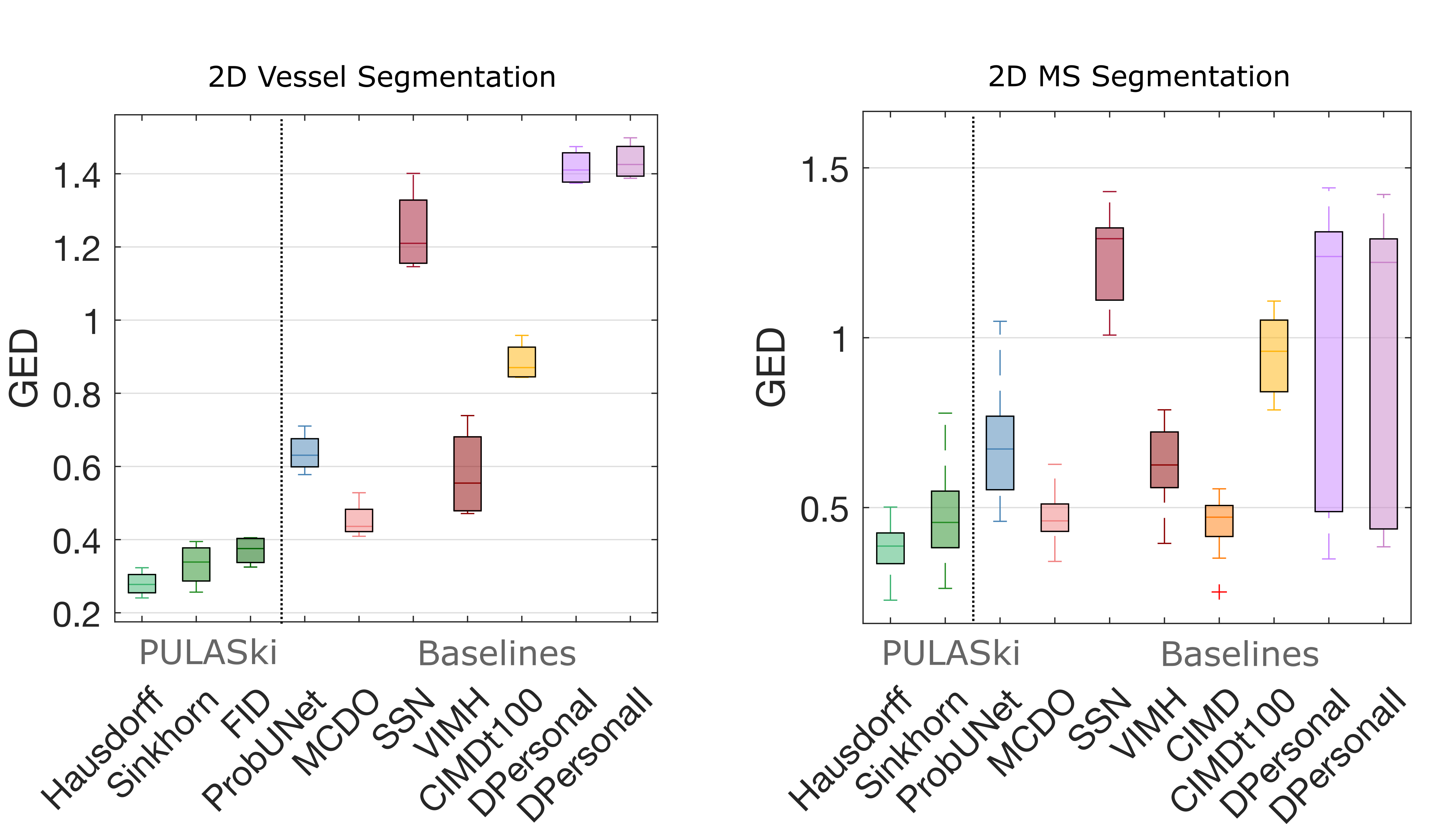}
  \caption{Quantitative assessment of distribution of generated segmentations per image compared to available data. Boxplots show variation in Generalised Energy Distance (GED) scores per image in the test set for all baselines and PULASKi with different statistical distances.}
  \label{suppFig:GED_supp}
\end{figure}

\begin{table}[!h]
\centering
\caption{Quantitative assessment of the distribution of generated segmentations using Krippendorff's alpha, calculated on the entire volume ($\mathrm{K\alpha_{all}}$) and only on voxels classified as vessel or lesion in at least one annotation ($\mathrm{K\alpha_{ROI}}$). 
 Values are indicated as mean over all images $\pm$ one standard deviation.  }
\label{supptab:kalpha}
\begin{tabular}{@{}lcc@{}}
\toprule
\multicolumn{3}{c}{\textbf{Vessel Segmentation}}   \\ \midrule
&  \multicolumn{2}{c}{2D} \\  
& $\mathrm{K\alpha_{all}}$ & $\mathrm{K\alpha_{ROI}}$ \\ \midrule
Annotation & $40.24 \pm 8.62$  & $22.35 \pm 10.9$ \\ \midrule
PULASki HD & $50.34 \pm 2.31$ & $31.28 \pm 2.39$ \\
PULASki SH & $57.62 \pm 1.18$ & $34.96 \pm 1.3$ \\
PULASki FID & $50.21 \pm 0.39$ & $29.66 \pm 0.61$ \\
Prob UNet & $99.83 \pm 0.01$ & $63.08 \pm 0.59$ \\
MCDO UNet & $90.00 \pm 1.36$ & $58.71 \pm 1.07$ \\
SSN & $100.0 \pm 0$ & $100.0 \pm 0$ \\
VIMH & $88.78 \pm 1.79$ & $57.90 \pm 1.22$ \\
CIMD ($T=100$) & $0.004 \pm 0.0012$  & $-0.94 \pm 0.0008$ \\
DPersonaI & $89.33 \pm 0.41$  & $51.94 \pm 3.15$ \\
DPersonaII & $94.34 \pm 0.49$  & $63.67 \pm 3.95$ \\ \midrule
&  \multicolumn{2}{c}{3D} \\ 
& $\mathrm{K\alpha_{all}}$ & $\mathrm{K\alpha_{ROI}}$  \\ \midrule
Annotation &  $40.24 \pm 8.62$ & $22.35 \pm 10.9$  \\ \midrule
PULASki HD  & $76.70 \pm 3.05$ & $50.56 \pm 3.18$ \\
PULASki SH & $82.85 \pm 2.27$ &  $53.30 \pm 2.36$ \\
Prob UNet  & $99.72 \pm 0.02$ & $62.74 \pm 0.17$ \\
MCDO UNet & $95.91 \pm 0.55$ &  $62.09  \pm 0.70$  \\
\toprule \bottomrule
\multicolumn{3}{c}{\textbf{MS Segmentation}}\\ \midrule
&  \multicolumn{2}{c}{2D} \\
& $\mathrm{K\alpha_{all}}$ & $\mathrm{K\alpha_{ROI}}$   \\ \midrule
Annotation & $60.40 \pm 14.06$ & $27.60 \pm 10.13$  \\ \midrule
PULASki HD & $62.06 \pm 11.88$  & $25.52 \pm 9.13$ \\
PULASki SH & $76.41 \pm 10.16$ & $40.75 \pm 8.85$  \\
Prob UNet & $99.93 \pm 0.03$ & $57.88 \pm 3.70$ \\
MCDO UNet & $82.93 \pm 8.19$  & $46.74 \pm 6.01$ \\
SSN & $100.0 \pm 0$ & $100.0 \pm 0$  \\
VIMH & $83.53 \pm 14.07$  & $45.78 \pm 10.13$  \\
CIMD ($T=1000$) & $36.54 \pm 18.48$  & $11.65 \pm 13.92$ \\
CIMD ($T=100$) & $0.002 \pm 0.007$  & $-0.8 \pm 0.01$ \\
DPersonaI & $45.35 \pm 32.03$  & $30.10 \pm 25.66$ \\
DPersonaII & $47.05 \pm 31.42$  & $28.95 \pm 21.64$ \\ \midrule
&  \multicolumn{2}{c}{3D} \\ 
& $\mathrm{K\alpha_{all}}$ & $\mathrm{K\alpha_{ROI}}$  \\ \midrule
Annotation & $60.40 \pm 14.06$ & $27.60 \pm 10.13$  \\ \midrule
PULASki HD & $86.58 \pm 5.81$ & $49.00 \pm 4.31$  \\
Prob UNet & $99.92 \pm 0.03$ & $58.39 \pm 2.50$  \\
MCDO UNet  & $95.87 \pm 2.32$ &  $55.47 \pm 1.99$ \\
\bottomrule
\end{tabular}
\end{table}

\end{document}